\documentclass{article}

\usepackage{microtype}
\usepackage{graphicx}
\usepackage{subcaption}
\usepackage{booktabs} % for professional tables
\usepackage{xurl}
\usepackage{pifont}

\usepackage{multirow}

\usepackage{threeparttable}
\usepackage{colortbl}
\usepackage{hyperref}
\usepackage{makecell}
\usepackage{anyfontsize}

\usepackage[accepted]{icml2024}

\usepackage{amsmath}
\usepackage{amssymb}
\usepackage{mathtools}
\usepackage{amsthm}
\usepackage{bm}
\usepackage{xspace}
\usepackage{enumitem}

\usepackage[capitalize,noabbrev]{cleveref}

\theoremstyle{plain}
\newtheorem{theorem}{Theorem}[section]
\newtheorem{proposition}[theorem]{Proposition}
\newtheorem{lemma}[theorem]{Lemma}
\newtheorem{corollary}[theorem]{Corollary}
\theoremstyle{definition}
\newtheorem{definition}[theorem]{Definition}
\newtheorem{assumption}[theorem]{Assumption}
\theoremstyle{remark}
\newtheorem{remark}[theorem]{Remark}

\usepackage[textsize=tiny]{todonotes}

\definecolor{mygray}{gray}{.9}

\newcommand{\project}{\mbox{{\textsc{BadPart}}}\xspace}

\newcommand{\mycomment}[1]{}
\newcommand{\TODO}[1]{\textcolor{teal}{[#1]}}

\newcommand{\zc}[1]{\textcolor{brown}{[#1]}}
\newcommand{\xz}[1]{\textcolor{red}{[XZ: #1]}}
\newcommand{\mingjie}[1]{\textcolor{orange}{[Mingjie: #1]}}
\newcommand{\gt}[1]{\textcolor{blue}{[GT: #1]}}
\newcommand{\sw}[1]{\textcolor{cyan}{[SF: #1]}}
\newcommand{\mc}[1]{\textcolor{green}{}}

\newcommand\redout{\bgroup\markoverwith{\textcolor{red}{\rule[0.5ex]{1pt}{1.5pt}}}\ULon}

\newcommand{\removed}[1]{\textcolor{red}{\sout{#1}}}

\newif\ifclean			  
\ifclean
	\renewcommand{\TODO}[1]{}	
	\renewcommand{\mingjie}[1]{}
	\renewcommand{\xz}[1]{}
	\renewcommand{\zc}[1]{}
        \renewcommand{\sw}[1]{}

	\renewcommand{\redout}[1]{}
	\renewcommand{\gt}[1]{}
	\renewcommand{\removed}[1]{}
	
\fi

\icmltitlerunning{BadPart: Unified Black-box Adversarial Patch Attacks}

\begin{document}

\twocolumn[
\icmltitle{BadPart: Unified Black-box Adversarial Patch Attacks \\
against Pixel-wise Regression Tasks}



\icmlsetsymbol{equal}{*}

\begin{icmlauthorlist}
\icmlauthor{Zhiyuan Cheng}{purdue}
\icmlauthor{Zhaoyi Liu}{scu}
\icmlauthor{Tengda Guo}{scu}
\icmlauthor{Shiwei Feng}{purdue}
\icmlauthor{Dongfang Liu}{rit}
\icmlauthor{Mingjie Tang}{scu}
\icmlauthor{Xiangyu Zhang}{purdue}
\end{icmlauthorlist}

\icmlaffiliation{purdue}{Department of Computer Science, Purdue University, West Lafayette, USA}
\icmlaffiliation{scu}{College of Computer Science, Sichuan University, Chengdu, China}
\icmlaffiliation{rit}{Department of Computer Engineering, Rochester Institute of Technology, Rochester, USA}

\icmlcorrespondingauthor{Zhiyuan Cheng}{cheng443@purdue.edu}
\icmlcorrespondingauthor{Xiangyu Zhang}{xyzhang@cs.purdue.edu}
\icmlcorrespondingauthor{Mingjie Tang}{tangrock@gmail.com}

\icmlkeywords{Machine Learning, ICML}

\vskip 0.3in
]


\printAffiliationsAndNotice{}  % leave blank if no need to mention equal contribution

\begin{abstract}

Pixel-wise regression tasks (e.g., monocular depth estimation (MDE) and optical flow estimation (OFE)) have been widely involved in our daily life in applications like autonomous driving, augmented reality and video composition. Although certain applications are security-critical or bear societal significance, the adversarial robustness of such models are not sufficiently studied, especially in the black-box scenario. In this work, we introduce the first unified black-box adversarial patch attack framework against pixel-wise regression tasks, aiming to identify the vulnerabilities of these models under query-based black-box attacks. We propose a novel square-based adversarial patch optimization framework and employ probabilistic square sampling and score-based gradient estimation techniques to generate the patch effectively and efficiently, overcoming the scalability problem of previous black-box patch attacks. Our attack prototype, named \project, is evaluated on both MDE and OFE tasks, utilizing a total of 7 models. \project surpasses 3 baseline methods in terms of both attack performance and efficiency. We also apply \project on the Google online service for portrait depth estimation, causing 43.5\% relative distance error with 50K queries. State-of-the-art (SOTA) countermeasures cannot defend our attack effectively.

\end{abstract}

\vspace{-15pt}
\section{Introduction}
\vspace{-5pt}
\begin{figure}
    \centering
    \includegraphics[width=0.9\columnwidth]{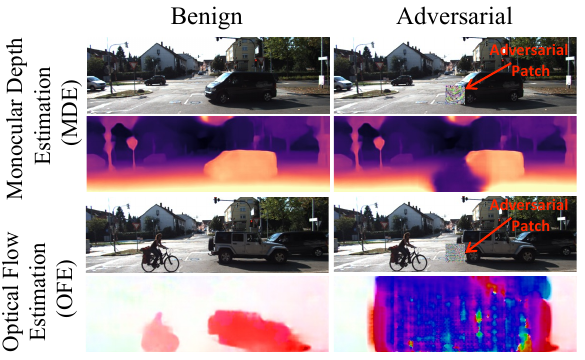}
    \vspace{-10pt}
    \caption{Adversarial patch attack on pixel-wise regression tasks.}
    \label{fig:highlight}
    \vspace{-18pt}
\end{figure}

Pixel-wise regression tasks represent a family of computer vision tasks that employ images as input and generate continuous regression values for each pixel in the input image. Examples of such tasks include monocular depth estimation (MDE), optical flow estimation (OFE), surface normal estimation (SNE), among others. The evolution of deep learning techniques has led to the development of numerous DNN models that have demonstrated impressive performance on these tasks, thereby enabling a variety of downstream applications such as autonomous driving, augmented reality, video composition, and more. Given the extensive and security-sensitive nature of some of these applications, it is crucial to examine the adversarial robustness of these tasks.

Previous adversarial patch attacks on pixel-wise regression models~\cite{cheng2022physical,cheng2023adversarial} have primarily been conducted in a white-box setting, where the attacker has complete knowledge of the target model and can use gradient-based optimization to create the patch and compromise the model.
However, this white-box assumption is not always feasible. For instance, models used in autonomous vehicles are typically closed-source~\cite{Tesla}. Moreover, some models are provided as an online service, preventing users from accessing their internals model structures~\cite{GoogleDepthApi}. In typical scenarios, black-box attacks are more realistic and could pose a greater threat to the security of these tasks. Existing black-box attack techniques are either based on training a substitute model or iteratively querying the model. In this study, we focus on the query-based approach as it reduces the cost of model training and only assumes that the attacker has access to the model output.
This leads us to a pivotal inquiry: ``\textit{can we craft an adversarial patch to compromise the pixel-wise regression model through iterative querying in a black-box manner?}" 

Answering the above question may bear direct relevance to applications of societal significance. 
For instance, in the realm of autonomous driving, attackers have been able to exploit the pixel-wise regression output of Tesla's depth estimation model through hacking~\cite{TeslaBreak}, and online services, such as Google Depth API~\cite{GoogleDepthApi} and Clipdrop API~\cite{ClipdropApi}, readily provide access to pixel-wise model outputs. The feasibility of such attacks within the context of autonomous driving would introduce a notable security concern. Meanwhile, on the other hand, the adversarial patch for online services could also act as a deterrent against unauthorized users who attempt to upload our photographs to those services for video composition.

To the best of our knowledge, we are the first to explore unified black-box patch attacks against pixel-wise regression tasks. Compared to classification tasks, pixel-wise regression tasks have denser (pixel-wise) query output and the resolution of input images is significantly higher. For example, the representative KITTI dataset~\cite{Uhrig2017THREEDV} used in MDE and OFE tasks includes images with a resolution of $1242\times 375$, while most previous black-box patch attacks on classification models~\cite{tao2023hard} use images from MINIST, CIFAR-10, GTSRB, etc., with a resolution less than $32\times 32$. The search space for a patch covering 1\% of the input images would expand exponentially as the resolution increases, hence the effectiveness and efficiency of prior methods are limited when adapted to our scenario.  

To address this issue and enhance the scalability of black-box patch generation, we leverage the domain knowledge of pixel-wise regression tasks and propose a novel square-based adversarial patch optimization framework. Specifically, in each iteration, we consider a small square area within the patch region to reduce the potential search space, then we introduce a batch of random noises on the square area to estimate the gradients of the region by evaluating the score of each noise in terms of attack performance. We sample the location of the square area probabilistically based on the pixel-wise error distribution, and propose novel score adjustment procedures for more precise gradient estimation. The source code is available at \url{https://github.com/Bob-cheng/BadPart}. Figure~\ref{fig:highlight} presents our attack performance. In summary, our contributions are as follows:
\vspace{-7pt}
\begin{itemize}[leftmargin=10pt]
\setlength\itemsep{-2pt}
    \item We introduce the first unified black-box adversarial patch attack framework against pixel-wise regression tasks (e.g., monocular depth estimation and optical flow estimation).
    \item We devise a square-based universal adversarial patch generation approach, employing probabilistic square sampling and score-based gradient estimation, to facilitate scalable black-box patch optimization. 
    \item We implement an attack prototype called \project (\underline{B}lack-box \underline{ad}versarial \underline{p}atch \underline{a}ttack against pixel-wise \underline{r}egression \underline{t}asks). We evaluate the attack performance of \project on both MDE and OFE tasks, utilizing a total of 7 models that encompass both popular and SOTA ones. Compared with three baseline methods that employ varying black-box optimization strategies, \project surpasses them in terms of both attack performance and efficiency. We also apply \project to attack a Google online service for portrait depth estimation~\cite{GoogleDepthApi}, resulting in 43.5\% relative distance error with 50K queries.
\end{itemize}
\vspace{-7pt}

% 1. Introduce pixel-wise regression models and applications. Then bring about the security concerns of it. 

% 2. Briefly introduce white-box attacks and bring about black-box attacks and our threat model.

% 3. Limitations of prior black-box attack approaches and our methods and designs to solve such problems and how we leverage domain knowledge.

\vspace{-5pt}
\section{Background and Related Work}
% \vspace{-5pt}

\textbf{Pixel-wise Regression Tasks.} Pixel-wise regression tasks generate continuous values for each input pixel, differing from pixel-wise classification tasks like semantic segmentation~\cite{wang2022learning,liang2023clustseg,liang2024clusterfomer,liu2021sg}, which assign a discrete class label to every pixel. Representative tasks of this type include monocular depth estimation (MDE)~\cite{moon2019parsimonious,watson2019self,wang2023planedepth}, optical flow estimation (OFE)~\cite{teed2020raft,ilg2017flownet,lu2023transflow}, and surface normal estimation (SNE)~\cite{zeng2019deep,lenssen2020deep,bae2021estimating}. In MDE models, the output comprises the estimated pixel-wise distance between the 3D scenario and the camera capturing the input image, with each pixel corresponding to a distance estimation. OFE models use two consecutive image frames as input and output the estimated motion of pixels (the ``optical flow") between the two frames. For each pixel in the first frame, a 2D vector is estimated, indicating its offset to the corresponding pixel in the second frame. SNE models output the estimated orientations of the surfaces in the input image, described with the "normal vectors" in 3D space that are perpendicular to the surface at the locations of pixels. Given that the first two tasks (MDE and OFE) have broader and more security-critical applications, such as autonomous driving~\cite{tesla-self-supervised}, visual SLAM~\cite{wimbauer2021monorec}, video composition~\cite{liew2023magicedit}, and augmented reality~\cite{bang2017camera}, our discussion primarily focuses on these two tasks. However, our proposed attack is a unified approach and can be readily applied to other pixel-wise regression tasks like SNE.

\textbf{Black-box Adversarial Attacks.} 
Existing black-box attacks can be broadly classified into two categories: substitute model-based attacks and query-based attacks. In the former, attackers construct a substitute model to execute white-box attacks and transfer the generated adversarial example to attack the victim model ~\cite{gao2020patch,liu2016delving,he2021drmi}. To construct the substitute model, attackers employ the same training set as the victim model or reverse-engineer/synthesize a similar dataset. Many works propose innovative training approaches to further improve the transferability~\cite{wu2020skip,feng2022boosting,wang2021enhancing,wang2021feature}. In the latter category of attacks, known as query-based attacks, attackers directly optimize the adversarial example by iteratively querying the victim model. Most black-box attacks focus on classification tasks, and they can be further divided into two groups: hard-label attacks and soft-label attacks, depending on the query output that the attacker can access. Hard-label attacks~\cite{chen2020hopskipjumpattack,li2020qeba,yan2020policy,tao2023hard} assume that the attacker can only access the predicted label of the victim model, while soft-label attacks~\cite{croce2022sparse,ilyas2018black,moon2019parsimonious} assume the prediction score of each class is available. With iterative queries, some prior works optimize the adversarial noise via gradient estimation~\cite{chen2019zo,zhang2021progressive,tao2023hard}, and some others rely on heuristic random search~\cite{croce2022sparse,andriushchenko2020square,duan2021adversarial}. Additionally, there are studies utilizing genetic algorithms to optimize the noise~\cite{ilie2021evoba,alzantot2019genattack}. Our work also falls under the category of query-based black-box attacks. However, we are the first to target pixel-wise regression tasks. We confront the domain-specific obstacle of high-resolution patch optimization, given that the SOTA black-box patch attack~\cite{tao2023hard} primarily concentrates on smaller patches and exhibits limited scalability in our scenario.

\vspace{-5pt}
\section{Problem Formalization}
% \vspace{-5pt}
The objective of the attack is to generate an adversarial patch, denoted as $\mathbf{p}$, for a black-box pixel-wise regression model $\mathcal{M}$. The desired result is that, irrespective of the input image $\mathbf{x}$, the attachment of the patch at location $\mathbf{q}$ on $\mathbf{x}$ will substantially degrade the model performance. Attackers can only query $\mathcal{M}$ for pixel-wise output. This black-box setting is highly practical as there are online services~\cite{GoogleDepthApi,ClipdropApi} that only allow users to upload custom images via API and return the depth estimation result. Additionally, in autonomous driving, attackers have shown to be able to reverse-engineer Tesla Autopilot to access the estimated depth map~\cite{TeslaBreak}. Formally, the optimization problem is expressed as follows: 
\begin{align}
    \max\limits_{\mathbf{p}}~ &Mean\left(\mathcal{F}\left(\mathcal{M}([\mathbf{x'}]_n) - \mathcal{M}([\mathbf{x_0'}]_n)\right)\right) \label{eq:prob_form}\\
    \text{\textbf{s.t.}}~&\mathbf{p}\in[0,1]^{3\times h\times h},\\
    \text{\textbf{where}}~&[\mathbf{x'}]_n = \Lambda([\mathbf{x}]_n, \mathbf{p}, \mathbf{q}), \label{eq:adv_img}\\
    &[\mathbf{x_0'}]_n = \Lambda([\mathbf{x}]_n, \mathbf{p_0}, \mathbf{q}), \label{eq:ori_img}\\
    &\mathcal{F}:\mathbb{R}^{n\times d\times H\times W} \rightarrow \mathbb{R}^{H\times W}\label{eq:error_func}\\
    &\mathbf{x} \in [0,1]^{3\times H\times W}, ~\mathbf{p}_0=\{0\}^{3\times h\times h},\\
    &\mathbf{q}\in \{(i,j)| i=\frac{h}{2}...H_{\!}-_{\!}\frac{h}{2}, j=\frac{h}{2}...W_{\!}-_{\!}\frac{h}{2}\}.
\end{align}
Here, $H$ and $W$ denote the height and width of the input image $\mathbf{x}$, $h$ the size of the patch $\mathbf{p}$, $n$ the number of images in the test set and  $\mathbf{q}$ the coordinates of the patch's center in $\mathbf{x}$. We use $\Lambda(\mathbf{a_1}, \mathbf{a_2}, \mathbf{a_3})$ to denote the process of attaching $\mathbf{a_2}$ to $\mathbf{a_1}$ at location $\mathbf{a_3}$, hence $[\mathbf{x'}]_n$ in Equation~\ref{eq:adv_img} refers to the image set attached with adversarial patch $\mathbf{p}$ at location $\mathbf{q}$, and $[\mathbf{x_0'}]_n$ in Equation~\ref{eq:ori_img} denotes the corresponding images attached with a black patch $\mathbf{p}_0$ (not optimized) as reference.
The output of model $\mathcal{M}$ has a dimension of $n\times d\times H\times W$, where $d$ refers to the output channels for each image. For MDE models $d$ equals 1 as the output is the estimated distance for each pixel, and for OFE models $d$ equals 2 since the model outputs the estimated pixel-wise offset vector (two dimensions). 
$\mathcal{F}$ in Equation~\ref{eq:error_func} denotes the function calculating the pixel-wise prediction error caused by the generated adversarial perturbation. 
\begin{align}
    \mathcal{F}(\mathbf{D})[i, j] &= Mean(\{\mathbf{\mathbf{D}}[k, 1, i, j]\mid k=1...n\})\label{eq:mde_f}\\
    \mathcal{F}(\mathbf{D})[i, j] &= Mean(\{\|\mathbf{D}[k, :, i, j]\|_2\mid k=1...n\})\label{eq:ofe_f}
\end{align}
Our attack goals are to maximize the estimated pixel-wise distances for MDE models or the $l_2$ norm of offset vectors for OFE models, hence, let $\mathbf{D}=\mathcal{M}([\mathbf{x'}]_n) - \mathcal{M}([\mathbf{x_0'}]_n)$, $\mathcal{F}$ is defined in Equation~\ref{eq:mde_f} for MDE models and in Equation~\ref{eq:ofe_f} for OFE models. For simplicity, see Equation~\ref{eq:pix_wis_error}, we use $\mathcal{F}_e([\mathbf{x'}]_n) \in \mathrm{R}^{H\times W}$ to denote the pixel-wise error caused by adversarial images $[\mathbf{x'}]_n$ in the following text. 
\begin{gather}
    \mathcal{F}_e([\mathbf{x'}]_n) \equiv \mathcal{F}\left(\mathcal{M}([\mathbf{x'}]_n) - \mathcal{M}([\mathbf{x_0'}]_n)\right)\label{eq:pix_wis_error}
\end{gather}

\begin{figure}[t]
    \centering
    \includegraphics[width=0.97\columnwidth]{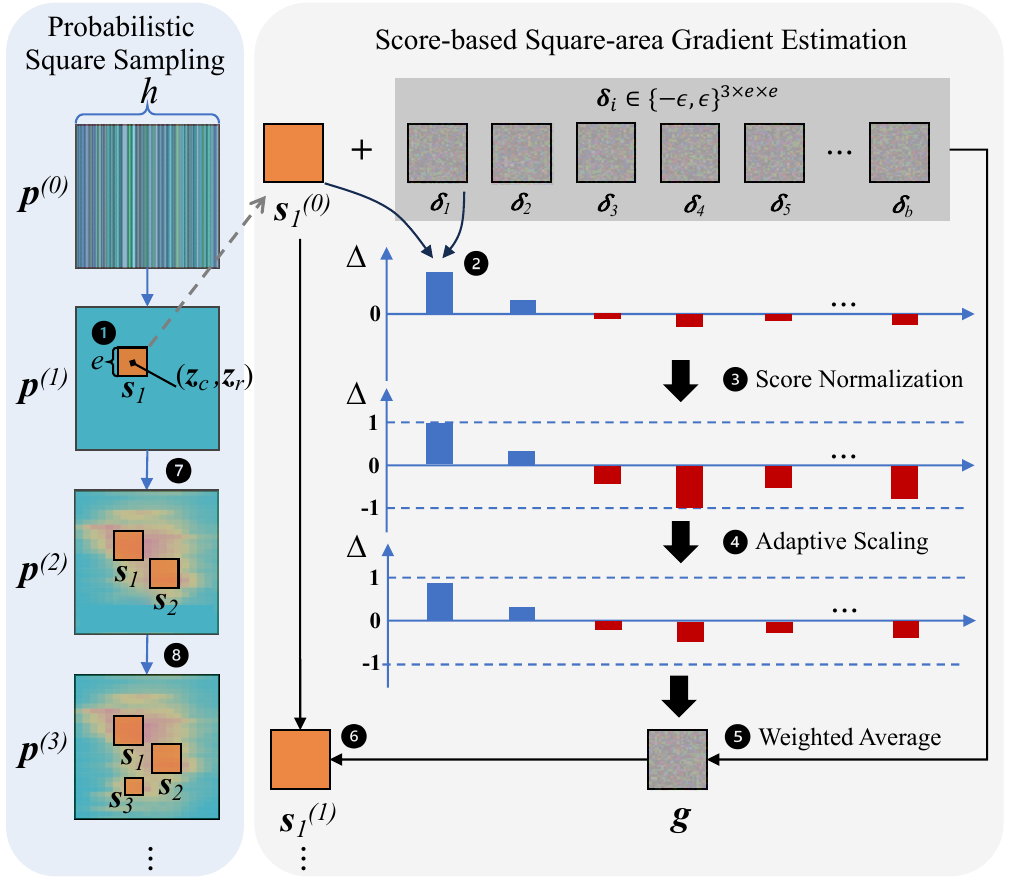}
    \vspace{-10pt}
    \caption{Overview of \project. }
    \label{fig:overview}
    \vspace{-10pt}
\end{figure}

\vspace{-10pt}
\section{Methods}\label{sec:method}

In this section, we first introduce our proposed square-based adversarial patch generation framework, as detailed in \S\ref{sec:framework} and explicated in Alg.~\ref{alg:BadPartAttack}. We then examine the two main components of the framework: the probabilistic square sampling, which is elucidated in \S\ref{sec:square_sampling} and Alg.~\ref{alg:square_sample}, and the score-based gradient estimation, presented in \S\ref{sec:grad_est} and Alg.~\ref{alg:GetGrad}.

% \vspace{-8pt}
\subsection{Square-based Patch Generation Framework}\label{sec:framework}
\vspace{-5pt}

The principal concept underlying our approach involves the iterative optimization of a square-shaped sub-area in the patch region, while altering the location and size of the target square area dynamically. Figure~\ref{fig:overview} illustrates the overview of the attack framework. The rectangles in teal ($\mathbf{p}^{(i)}$) on the left of Figure~\ref{fig:overview} denote the patch region at different optimization stages, and the small squares in orange ($\mathbf{s}_i$) refer to the sampled sub-areas to optimize. Steps \ding{182}, \ding{188} and \ding{189} denote the selection of the square areas, and steps \ding{183}-\ding{187} present the procedure of optimizing a selected square. These selection and optimization are carried out alternately. The strategy of selecting a square area within the broader patch region serves to effectively constrain the large search space inherent to the entire patch.
It is inspired from 
SquareAttack~\cite{andriushchenko2020square} which attacks classification models via random search and leverages square areas as the perturbation units.$_{\!}$ The$_{\!}$ rationale$_{\!}$ for$_{\!}$ favoring$_{\!}$ a$_{\!}$ square$_{\!}$ shape$_{\!}$ is$_{\!}$ that$_{\!}$ modern$_{\!}$ image-processing$_{\!}$ models$_{\!}$ predominantly$_{\!}$ utilize$_{\!}$ convolutional$_{\!}$ layers$_{\!}$ for$_{\!}$ feature$_{\!}$ extraction.$_{\!}$ The$_{\!}$ filter$_{\!}$ kernels$_{\!}$ within$_{\!}$ these$_{\!}$ layers$_{\!}$ are$_{\!}$ inherently$_{\!}$ square-shaped,$_{\!}$ thereby$_{\!}$ making$_{\!}$ the$_{\!}$ square$_{\!}$ setting$_{\!}$ as$_{\!}$ the$_{\!}$ most$_{\!}$ efficient
~\cite{andriushchenko2020square}. 
Unlike SquareAttack, \project iteratively updates each square area using novel gradient estimation rather than a single-step trial. This refinement transforms our approach into a more precise optimization process rather than random search, thereby strengthening the attack's effectiveness. Furthermore, we focus on pixel-wise regression tasks and utilize the domain knowledge of pixel error distribution to probabilistically select square locations, significantly boosting the efficiency.

Alg.~\ref{alg:BadPartAttack} describes the proposed universal adversarial patch generation framework. 
To begin with, we initialize the patch region with vertical strips, where the color of each stripe is sampled uniformly at random from $\{0, 1\}^3$ (see $\mathbf{p}^{(0)}$ in Figure~\ref{fig:overview} and line~4 in Alg.~\ref{alg:BadPartAttack}). Then we attach the patch to the validation images at location $\mathbf{q}$, and record the overall error $\omega^*$ as the initialized best attack performance caused by the perturbations (lines~5-6). After initialization, we start the iterations of square sampling. In each iteration, we first calculate the pixel-wise error map $\mathbf{M}$ caused by the latest patch on validation images (lines~9-10), then, in lines 11-12, we call the probabilistic square sampling algorithm (Alg.~\ref{alg:square_sample} explained in \S\ref{sec:square_sampling}) with $\mathbf{M}$ and the iteration index as input, getting the sampled square area (step \ding{182} in Figure~\ref{fig:overview}). Next, we start optimizing the square area. In each round of optimization, we first estimate the gradients of the square area on a random training image (lines~15-16). Details of the score-based gradient estimation (steps \ding{183}-\ding{186} in Figure~\ref{fig:overview}) will be explained in \S\ref{sec:grad_est} and Alg.~\ref{alg:GetGrad}. Then we update the square area with Adam optimizer using the estimated gradients (step~\ding{187}), and evaluate the attack performance $\omega$ of the latest patch (lines~17-19). We update the best performance $\omega^*$ to $\omega$ if $\omega$ is better and continue the next step of optimization on this square. If the best performance is not updated for $T_1$ steps, which indicates this square is sufficiently explored, we stop optimizing this square and continue the$_{\!}$ next$_{\!}$ iteration$_{\!}$ of$_{\!}$ square$_{\!}$ sampling$_{\!}$ and$_{\!}$ optimization$_{\!}$ (lines$_{\!}$ 20-21).$_{\!}$ Line$_{\!}$ 7$_{\!}$ and$_{\!}$ 23$_{\!}$ are$_{\!}$ noise$_{\!}$ bound-related,$_{\!}$ which$_{\!}$ will$_{\!}$ be$_{\!}$ explained$_{\!}$ in$_{\!}$ \S\ref{sec:grad_est}.

\begin{figure}[t]
\vspace{-10pt}
\begin{algorithm}[H]
   \caption{Square-based patch generation framework}
   \label{alg:BadPartAttack}
\begin{algorithmic}[1]
\fontsize{8.5}{11}\selectfont
   \STATE {\bfseries Input:} Pixel-wise Error Function $\mathcal{F}_e$, Training Images $[\mathbf{x^t}]_m$, Validation Images $[\mathbf{x^v}]_n$, Patch Location $\mathbf{q}$, Patch Size $h$.
   \STATE {\bfseries Output:} The generated patch $\mathbf{p}$.
   \FUNCTION{\textsc{BadPart}($\mathcal{F}_e$, $[\mathbf{x^t}]_m$, $[\mathbf{x^v}]_n$, $\mathbf{q}$, $h$)}
        \STATE $\mathbf{p}$ $\gets$ Initialize a patch $\mathbf{p} \in [0, 1]^{3\times h\times h}$ with vertical strips. 
        \STATE $[\mathbf{x'^v}]_n \gets \Lambda([\mathbf{x^v}]_n, \mathbf{p}, \mathbf{q})$
        \COMMENT{{\color{blue} $\rhd$ Attach patch $\mathbf{p}$ to $[\mathbf{x^v}]_n$}}
        \STATE $\omega^* \gets Mean(\mathcal{F}_e([\mathbf{x'^v}]_n))$
        \COMMENT{{\color{blue} $\rhd$ Record largest error as $\omega^*$}}
        \STATE $\epsilon \gets \alpha$
        \COMMENT{{\color{blue} $\rhd$ Initialize noise bound $\epsilon$ as $\alpha$}}
        \FOR{$iter$ in $0 ... max\_iters$}
           \STATE $[\mathbf{x'^v}]_n \gets \Lambda([\mathbf{x^v}]_n, \mathbf{p}, \mathbf{q})$
           \STATE $\mathbf{M} \gets \mathcal{F}_e([\mathbf{x'^v}]_n )$
           \STATE $\mathbf{z}, e \gets \textsc{GetSquareArea}(iter, \mathbf{M}, \mathbf{q}, h)$  
           \COMMENT{{\color{blue}  $\rhd$ Alg.~\ref{alg:square_sample}}}
           \STATE $\mathbf{s} \gets \mathbf{p}\left[\mathbf{z}_c - \frac{e}{2}...\mathbf{z}_c + \frac{e}{2},~~\mathbf{z}_r -\frac{e}{2}...\mathbf{z}_r + \frac{e}{2}\right]$
           \STATE /*** \textbf{Optimize the square area} ***/
           \FOR{$step$ in $0 ... max\_steps$}
                \STATE $\mathbf{x} \gets$ Randomly sample an image from $[\mathbf{x^t}]_m$.
               \STATE $\mathbf{g} \gets \textsc{GetGrad}(\mathcal{F}_e, \mathbf{x}, \mathbf{p}, \mathbf{q}, \mathbf{z}, e, \epsilon)$
               \COMMENT{{\color{blue}  $\rhd$ Alg.~\ref{alg:GetGrad}}}
               \STATE $\mathbf{s} \gets Optimizer(\mathbf{s}, \mathbf{g})$
               % \STATE $\mathbf{p}\left[i - \frac{e}{2}...i + \frac{e}{2},~~j -\frac{e}{2}...j + \frac{e}{2}\right] \gets \mathbf{s}$ 
               \STATE $\mathbf{p} \gets \Lambda(\mathbf{p}, \mathbf{s}, \mathbf{z})$
               \COMMENT{{\color{blue} $\rhd$ Attach $\mathbf{s}$ to $\mathbf{p}$ at location $\mathbf{z}$}}
               % \IF {$step~\%~\beta = 0 $}
                    \STATE $\omega \gets Mean \left(\mathcal{F}_e\left(\Lambda\left([\mathbf{x^v}]_n, \mathbf{p}, \mathbf{q}\right)\right)\right)$
                    \STATE \textbf{if} $\omega > \omega^*$ \textbf{then} $\omega^* \gets \omega$;
                    \STATE \textbf{if} $\omega^*$ is not updated for $T_1$ steps \textbf{then} break loop;
               % \ENDIF \COMMENT{{\color{blue} $\rhd$ Validate $\mathbf{p}$'s performance every $\beta$ steps}}
            \ENDFOR
            \STATE\textbf{if} {$\omega^*$ is not updated for $T_2$ iterations} \textbf{then} $\epsilon \gets \epsilon * \gamma$;
                % \COMMENT{{\color{blue} $\rhd$ Decay the noise bound with factor $\gamma$}}
           % \ENDIF
       \ENDFOR
       % \STATE \textbf{return} $\mathbf{p}$
   \ENDFUNCTION
\end{algorithmic}
% \vspace{-10pt}
\end{algorithm}
\vspace{-10pt}
\end{figure}
% \vspace{-10}

% \vspace{-8pt}
\subsection{Probabilistic Square Sampling}\label{sec:square_sampling}
\vspace{-5pt}

The sampling algorithm is designed to enhance the probability of selecting locations within the patch region that are more vulnerable to adversarial perturbations. 
As indicated in \cite{cheng2022physical} and \cite{cheng2023fusion}, the identification of vulnerable areas on images is critical for improving attack performance. To ascertain these vulnerable areas in the beginning, we employ an initialization phase ($K$ iterations), wherein we randomly sample the square location in a uniform manner. This period is denoted as $\mathbf{p^{(1)}}$ in Figure~\ref{fig:overview}.  After the initialization phase, the pixel-wise error map $\mathbf{M}$ caused by the latest patch is utilized as an indication of the vulnerable areas and we leverage this map to calculate the probability distribution of location sampling. Those areas with larger errors obtain higher probability. The background color of the patch $\mathbf{p^{(1)}}, \mathbf{p^{(2)}}$ and $\mathbf{p^{(3)}}$ in Figure~\ref{fig:overview} illustrate the sampling probability distribution of square locations. 

Alg.~\ref{alg:square_sample} describes the probabilistic square area sampling algorithm, whose output are the sampled square location in the patch region and the square size. In accordance with SquareAttack, the square size is obtained from a predefined schedule regarding the iteration index (line 4), which is detailed in Appendix~\ref{app:exp_setting}. As the index escalates, the size diminishes, indicative of a transition from coarse to fine-grained optimization (see $\mathbf{s_3}$ after step~\ding{189} in Figure~\ref{fig:overview}). The initialization phase of uniformly sampling lasts $K$ iterations (lines~5-6), after which we start error-based probabilistic sampling. We first smooth the pixel-wise error map $\mathbf{M}$ with a filter kernel that has the same size of the square, which is to avoid extreme values at certain locations (line~9). Then we crop out the patch region and normalize the smoothed error map to $[0, 1]$ followed by applying the softmax function to transform the error map into the sampling probabilities for different locations (lines~10-11). At last, see step~\ding{188} or \ding{189} in Figure~\ref{fig:overview}, the square location ($\mathbf{s_2}$ or $\mathbf{s_3}$) is sampled based on the probability distribution (line~12).

\begin{figure}[t]
\vspace{-10pt}
\begin{algorithm}[H]
   \caption{Probabilistic square area sampling}
   \label{alg:square_sample}
\begin{algorithmic}[1]
\fontsize{8.5}{11}\selectfont
   \STATE {\bfseries Input:} Iteration Index $iter$, Pixel-wise Error Map $\mathbf{M}$, Patch Location $\mathbf{q}$, Patch Size $h$.
   \STATE {\bfseries Output:} Square Location $\mathbf{z}$, Square Size~$e$.
   \FUNCTION{ $\textsc{GetSquareArea}(iter, \mathbf{M}, \mathbf{q}, h)$}
       \STATE $e \gets SizeSche(iter)$ \COMMENT{{\color{blue} $\rhd$ Use a pre-defined size schedule}}
       \IF{$iter \leq$ \text{Initialization period} $K$}
       \STATE $\mathbf{z} \gets $ Sample a location index from $\{0...h\}^2$ uniformly.
       \ELSE 
       \STATE \textbf{/*** Error-based probabilistic sampling ***/}
       \STATE $\mathbf{M} \gets $ Smooth error map $\mathbf{M}$ with kernel size $e \times e$.
       \STATE $\mathbf{M}_{\mathbf{q},h} \gets \mathbf{M}\left[\mathbf{q}_c-\frac{h}{2}...\mathbf{q}_c+\frac{h}{2},~~\mathbf{q}_r-\frac{h}{2}...\mathbf{q}_r+\frac{h}{2}\right]$
       \STATE $\mathbf{prob} \gets Softmax(\mathbf{M}_{\mathbf{q},h} / Max(\mathbf{M}_{\mathbf{q},h}))$
       \STATE $\mathbf{z} \gets$ Sample a location index from $\{0...h\}^2$ with the \\ ~~~~~~~~~probability distribution $\mathbf{prob}$.
       \ENDIF 
   \ENDFUNCTION
\end{algorithmic}
\end{algorithm}
\vspace{-10pt}
\end{figure}
% \vspace{-8pt}
\subsection{Score-based Gradient Estimation}\label{sec:grad_est}
\vspace{-5pt}
In this section, we introduce the score-based square-area gradient estimation method. 
As shown on the right side of Figure~\ref{fig:overview}, upon determination of the square area, we proceed to generate a batch of noise $[\bm{\delta}]_b$ within the confines of the square. This noise is constrained by a small threshold $\epsilon$, thereby facilitating the exploration of the adjacent high-dimensional space. Compared with the zeroth order optimization~\cite{chen2017zoo} that estimates the gradient pixel by pixel, our method is more efficient as the unit of gradient estimation is a square area. Values of the noise tensor are either $\epsilon$ or $-\epsilon$ as \cite{moon2019parsimonious} has indicated that the optimal adversarial noise is mostly found on vertices of the bound. Subsequently, for each instance of noise, we utilize the alteration in attack performance, consequent to the application of the noise on the current square area, as the evaluative score $\Delta$ of the noise (see step~\ding{183}. Negative scores denote negative impact on attack performance). Since the scores could be very small and imbalanced among the positive and negative ones, we normalize the positive and negative scores by scaling them to [0,1] and [0,-1] respectively (step~\ding{184}). Next, we conduct adaptive scaling to allocate greater weights to the side (positive or negative) with fewer elements (step~\ding{185}). This procedure is inspired from \cite{tao2023hard}, which subtracts the mean score from the positive and negative indicators. However, in our scenario of regression tasks, subtracting the mean score could potentially change the sign of scores, hence we opt to divide the positive (or negative) scores by the number of positive (or negative) elements for scaling purpose. Subsequently, we compute the weighted average of $[\mathbf{\delta}]_b$ by employing the scaled scores $[\Delta]_b$, followed by normalizing the output through dividing it by its $l_2$-norm. This procedure yields the estimated gradients (step~\ding{186}).

\begin{figure}[t]
\vspace{-10pt}
\begin{algorithm}[H]
   \caption{Score-based square-area gradient estimation}
   \label{alg:GetGrad}
\begin{algorithmic}[1]
\fontsize{8.5}{11}\selectfont
   \STATE {\bfseries Input:} Pixel-wise Error Function $\mathcal{F}_e$, Image $\mathbf{x}$, Patch~$\mathbf{p}$, Patch Location~$\mathbf{q}$, Square Location $\mathbf{z}$, Square Size $e$, Noise Bound $\epsilon$.
   \STATE {\bfseries Output:} Estimated Gradient $\mathbf{g}$.
   \FUNCTION{ $\textsc{GetGrad}(\mathcal{F}_e, \mathbf{x}, \mathbf{p}, \mathbf{q}, \mathbf{z}, e, \epsilon)$}
       % \STATE $\mathbf{x} \gets$ Randomly sample an image from $[\mathbf{x^t}]_m$.
       \STATE $\mathbf{x'} \gets \Lambda(\mathbf{x}, \mathbf{p}, \mathbf{q})$
       \COMMENT{{\color{blue} $\rhd$ Get image with current patch attached}}
       \STATE $\mathbf{s} \gets \mathbf{p}\left[\mathbf{z}_c - \frac{e}{2}...\mathbf{z}_c + \frac{e}{2},~~\mathbf{z}_r -\frac{e}{2}...\mathbf{z}_r + \frac{e}{2}\right]$
       \STATE \textbf{/*** Calculate noise scores ***/}
       \STATE $[\bm{\delta}]_b \gets $ Generate $b$ random noise $\bm{\delta} \in \{-\epsilon, \epsilon\}^{3\times e\times e}$.
       % \STATE $[\hat{\mathbf{s}}]_b \gets \mathbf{s} + [\bm{\delta}]_b$ 
       % \STATE $[\hat{\mathbf{p}}]_b \gets$ Add $[\bm{\delta}]_b$ onto $\mathbf{s}$ at location $\mathbf{z}$ of $\mathbf{p}$.
       \STATE $[\hat{\mathbf{p}}]_b \gets \Lambda(\mathbf{p}, \mathbf{s} + [\bm{\delta}]_b, \mathbf{z})$
       \COMMENT{{\color{blue} $\rhd$ Add $[\bm{\delta}]_b$ to $\mathbf{s}$ and attach to $\mathbf{p}$}}
       % \STATE $[\hat{\mathbf{p}}]_b \gets$ Apply $[\hat{\mathbf{s}}]_b$ to location $\mathbf{z}$ of $\mathbf{p}$.
        \STATE $[\hat{\mathbf{x}}]_b \gets \Lambda(\mathbf{x}, [\hat{\mathbf{p}}]_b, \mathbf{q})$ 
        \COMMENT{{\color{blue} $\rhd$ Get images with noise set applied}}
        \STATE $[\Delta]_b \gets 0$ 
        \COMMENT{{\color{blue} $\rhd$ Initialize score for each noise}}
        \FOR{$\mathbf{\hat{x}_i}$ in $[\hat{\mathbf{x}}]_b$ }
            \STATE $\Delta_i \gets Mean(\mathcal{F}_e(\mathbf{\hat{x}_i})) - Mean(\mathcal{F}_e(\mathbf{x'}))$
        \ENDFOR \COMMENT{{\color{blue} $\rhd$ Update $[\Delta]_b$}}
        \STATE \textbf{/*** Score Adjustment ***/}
        \STATE $[\Delta]^+\gets [\Delta]^+ / \operatorname{max}([\Delta]^+)$; $[\Delta]^-\gets [\Delta]^-/|\operatorname{min}([\Delta]^-)|$
        \STATE $[\Delta]^+ \gets [\Delta]^+ /\#([\Delta]^+)$; ~~$[\Delta]^- \gets [\Delta]^- /\#([\Delta]^-)$
        \STATE $\mathbf{g} \gets$ Weighted sum of noise $[\bm{\delta}]_b$ using weights $[\Delta]_b$.
        \STATE $\mathbf{g} \gets \sqrt{3\cdot e \cdot e} \cdot\mathbf{g} / \|\mathbf{g}\|_2  $
        \COMMENT{{\color{blue} $\rhd$ Return estimated gradient}}
   \ENDFUNCTION
\end{algorithmic}
\end{algorithm}
\vspace{-10pt}
\end{figure}

\begin{table*}[t]
    \centering
    \caption{\small The mean error (depth estimation error (DEE, unit: meters) for MDE models and end point error (EPE, unit: pixels) for OFE models) caused by \project and other baseline methods on different target models. Larger values denote better attack performance.}
    \label{tab:atk_perf}
\scalebox{0.77}{
\begin{threeparttable}
\definecolor{mycolor}{rgb}{.9,.9,.9} % Define a color
\begin{tabular}{c|cccc>{\columncolor{mycolor}}c|cccc>{\columncolor{mycolor}}c|cccc>{\columncolor{mycolor}}c}
\toprule
 & \multicolumn{5}{c}{57 $\times$ 57 Patch (1\%)} & \multicolumn{5}{c}{80 $\times$ 80 Patch (2\%)} & \multicolumn{5}{c}{100 $\times$ 100 Patch (3\%)} \\
 Models & GA & HB & P-RS & \textbf{Ours} & WB & GA & HB & P-RS & \textbf{Ours} & WB & GA & HB & P-RS & \textbf{Ours} & WB \\ \midrule\midrule
Monodepth2 & 0.05 & 7.76 & 56.83 & \textbf{79.29} & 89.23 & 0.01 & 19.64 & 78.47 & \textbf{89.75} & 90.99 & 0.61 & 18.03 & 89.02 & \textbf{91.11} & 91.38 \\
DepthHints & 0.30 & 2.25 & 21.21 & \textbf{71.14} & 89.47 & 0.76 & 3.92 & 55.34 &\textbf{70.38}  & 90.15 & 1.43 & 2.24  & 76.85 & \textbf{87.32} & 90.81 \\
SQLDepth & 0.03 & 0.06 & 28.63 & \textbf{48.74} & 55.09 & 0.43 & 0.16 & 39.89 & \textbf{54.14} & 61.05 & 0.30 & 0.45  & 48.66 & \textbf{54.51} & 61.91   \\
PlaneDepth & 0.61 & 0.83 & 4.07 & \textbf{48.22} & 90.11 & 1.71 & 1.07 & 7.67 & \textbf{46.85} & 90.24 & 1.84 & 1.47 & 26.58 & \textbf{80.07} & 82.62\\ \midrule
FlowNetC & 5.42 & 4.21 & 5.32 & \textbf{583.20} & 2403.32 & 4.08 & 3.55 & 447.60 & \textbf{1212.13} & 3585.12 & 5.15 & 4.43 & 640.30 & \textbf{1033.21} & 2345.54 \\
FlowNet2 & 2.24 & 12.30 & 2.64 &\textbf{30.72}  & 55.81 & 1.65 & 7.31 & 1.77 & \textbf{32.42} & 725.30 & 1.27 & 10.39 & 5.94 & \textbf{27.82} & 194.40 \\
PWC-Net & 1.93 &2.04  & 2.35 & \textbf{4.87} & 53.68 & 1.73 & 1.90 & 1.66 & \textbf{5.26} & 149.30 & 1.48 & 1.53 & 1.44 & \textbf{5.32} & 55.31 \\ \bottomrule
% RAFT &  &  &  &  &  &  &  &  &  &  &  &  &  &  &  \\ 
\end{tabular}
\begin{tablenotes}
\footnotesize
\item * Bold texts denote the best attack performance among black-box methods. Abbreviations. GA: GenAttack~\cite{alzantot2019genattack}, HB: HardBeat~\cite{tao2023hard}, P-RS: Patch-RS~\cite{croce2022sparse}, WB: White-box Attack for reference.
\end{tablenotes}
\end{threeparttable}
}
% \vspace{-10pt}
\end{table*}

Details of the algorithm can be found in Alg.~\ref{alg:GetGrad}. Lines~4-5 attach the latest patch onto the input image, and crop out the square area, where $\mathbf{z}_c$ and $\mathbf{z_r}$ denote the column and row index of the square center within the patch. Lines~7-9 generate the set of random noise, and apply them to the square area, creating candidate input images $[\hat{\mathbf{x}}]_b$. Lines~10-13 calculate the scores $[\Delta]_b$ by comparing the attack performance of $[\hat{\mathbf{x}}]_b$ with the reference one $\mathbf{x'}$. Subsequently, Lines~15-16 adjust the scores by normalization and adaptive scaling, and Lines~17-18 conduct the weighted average and normalization operations, achieving the final gradients. Note that, as shown in line~7 and 23 of Alg.~\ref{alg:BadPartAttack}, the threshold $\epsilon$ of the noise is initialized as $\alpha$ and will decay if the best attack performance $\omega^*$ is not updated for $T_2$ iterations of square selection. The decay factor $\gamma$ is set to 0.98 in our experiments.

% \vspace{-5pt}
\section{Evaluation}
% \vspace{-5pt}
In this section, we evaluate \project on 2 kinds of tasks including 7 subject models. We compare with 3 baseline black-box attack methods and a white-box one. A set of ablation studies are discussed and the source code is provided here: \url{https://github.com/Bob-cheng/BadPart}.

% \vspace{-8pt}
% \vspace{-5pt}
\subsection{Experiment Setup}\label{sec:eval_setup}

\begin{figure*}[t]
\centering
    \includegraphics[width=0.95\textwidth]{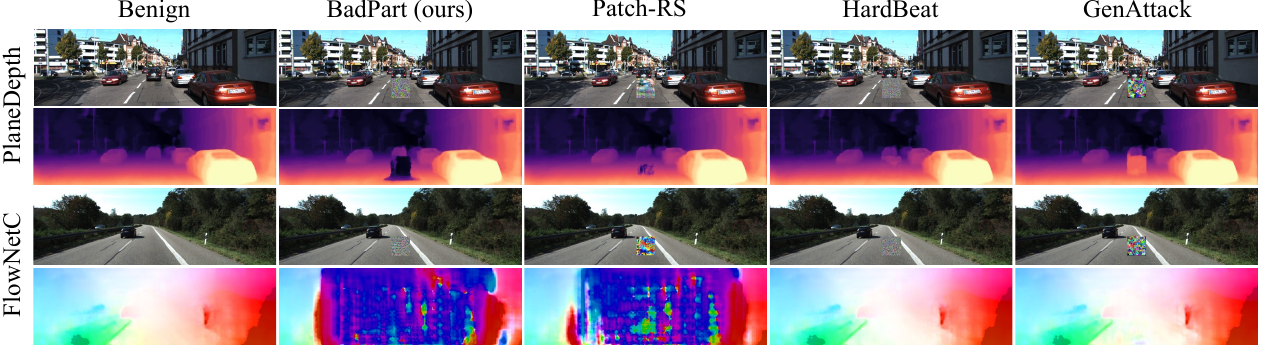}
    \vspace{-10pt}
        \caption{Examples of the qualitative attack performance of \project and the baselines.}
        \label{fig:vis_main_quali}
        % \vspace{-12pt}
\end{figure*}

\textbf{Tasks \& Models.} We evaluate \project on two pixel-wise regression tasks of MDE and OFE. For MDE, we use Monodepth2~\cite{godard2019digging}, DepthHints~\cite{watson2019self}, SQLDepth~\cite{wang2023sqldepth} and PlaneDepth~\cite{wang2023planedepth} as the target models. For OFE, we attack FlowNetC~\cite{dosovitskiy2015flownet}, FlowNet2~\cite{ilg2017flownet} and PWC-Net~\cite{sun2018pwc}. These models are selected since they cover both the popular and SOTA models, and we attack the publicly available models with the highest input resolutions from their official repositories. MDE models were trained on the KITTI depth prediction dataset~\cite{Uhrig2017THREEDV} and OFE models were trained on the KITTI flow 2015~\cite{Menze2015ISA}.

\begin{figure*}[t]
    \centering
    \begin{subfigure}[b]{0.23\textwidth}
        \includegraphics[width=\textwidth]{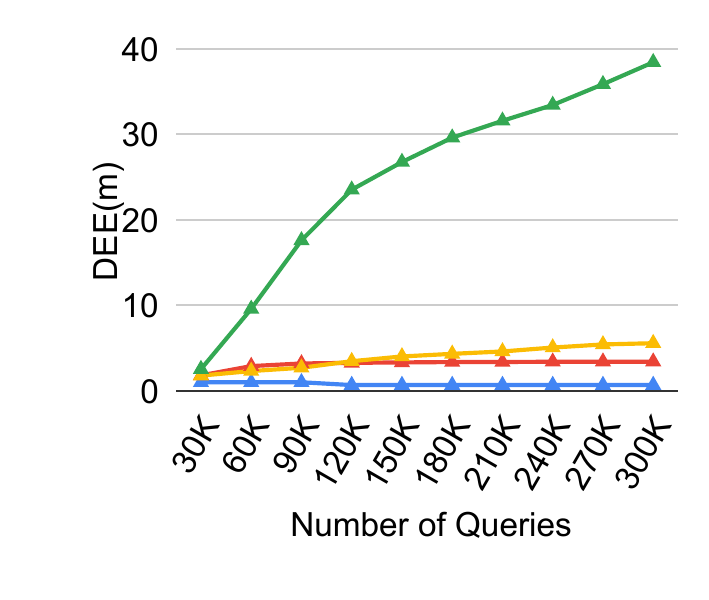}
        \caption{Depthhints}
        \label{fig:image1}
    \end{subfigure}
    \hfill
    \begin{subfigure}[b]{0.23\textwidth}
        \includegraphics[width=\textwidth]{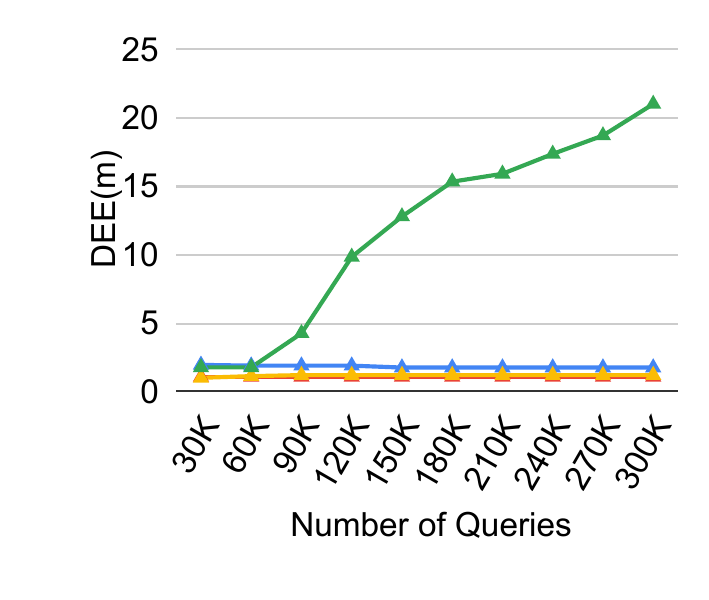}
        \caption{Planedepth}
        \label{fig:image2}
    \end{subfigure}
    \hfill
    \begin{subfigure}[b]{0.24\textwidth}
        \includegraphics[width=\textwidth]{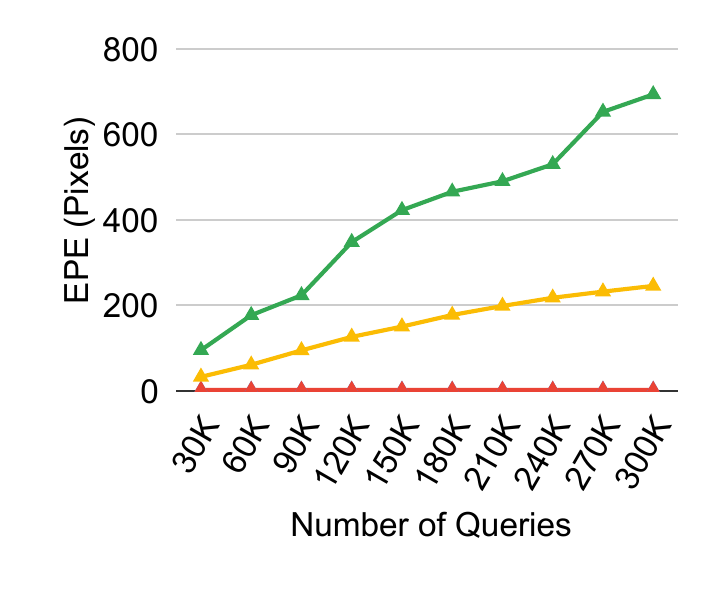}
        \caption{FlowNetC}
        \label{fig:image3}
    \end{subfigure}
    \hfill
    \begin{subfigure}[b]{0.275\textwidth}
        \includegraphics[width=\textwidth]{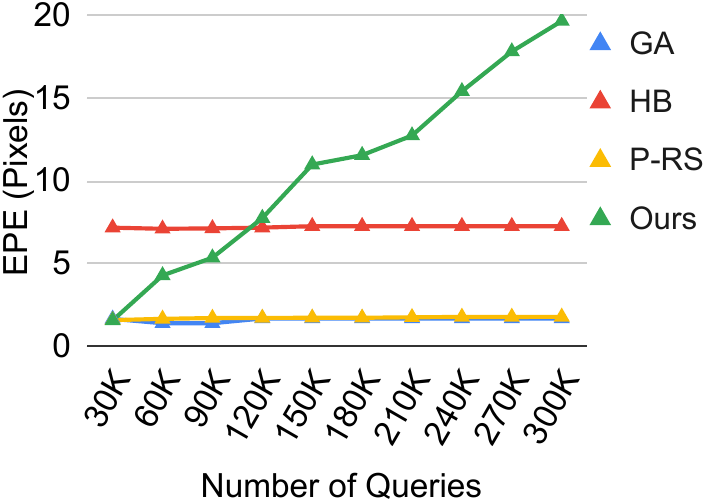}
        \caption{FlowNet2}
        \label{fig:image4}
    \end{subfigure}
    \vspace{-10pt}
    \caption{Comparison of query efficiency between \project and the baseline methods on four models (2\% patch).}
    \label{fig:query_effi}
    \vspace{-5pt}
\end{figure*}

\textbf{Baselines \& Metrics.} There are no direct baselines available due to a lack of previous research on black-box attacks against pixel-wise regression tasks. Hence we adapt three black-box patch attacks on image classification to our scenario as baselines. These include Patch-RS~\cite{croce2022sparse}, a SOTA soft-label attack that employs random search and is akin to the patch-attack variant of SquareAttack~\cite{andriushchenko2020square}; HardBeat~\cite{tao2023hard}, a SOTA hard-label attack that estimates gradients; and GenAttack~\cite{alzantot2019genattack}, a conventional approach using genetic algorithms.
We also compare \project with the white-box attack 
as reference. Discussion on more prior attacks can be found in Appendix~\ref{app:atk_discuss}. We employ the mean depth estimation error (DEE, unit: meters) for MDE models and the mean end point error (EPE, unit: pixels) for OFE models as metrics to denote attack performance (the higher the better), which are aligned with the metrics used in prior white-box attacks on the two tasks~\cite{cheng2022physical,ranjan2019attacking}. They are calculated on the patched area and reflect the error caused by the generated adversarial noise. 

\textbf{Attack Settings.} We use 100 scenes from KITTI flow dataset as our training set and another 5 scenes as the validation set during patch generation. (i.e., $m$ equals 100 and $n$ equals 5 in Alg.~\ref{alg:BadPartAttack}.) We evaluate the attack performance on a test set with 100 new scenes from the dataset, which covers different driving conditions (e.g., various road types, weathers and lighting). Each scene has two consecutive image frames since two images are required for the input of OFE models. For MDE models, we only feed the first frame. Detailed attack settings can be found in Appendix~\ref{app:exp_setting}, including hyper-parameter settings, devices, runtime overhead, etc.

% \vspace{-8pt}
% \vspace{-5pt}
\subsection{Attack Performance}
We compare the attack performance of \project with other baseline methods. We report the maximum attack performance of each method until convergence or after 1000K queries, whichever first. We also include the performance of white-box attacks as references. Three different patch sizes are evaluated and the patch locates at the center of the image. 
Table~\ref{tab:atk_perf} reports the result. The first column denotes various pixel-wise regression models under attack, and the following columns represent attack performance of different methods and the white-box reference. As shown, \project obtains the best attack performance on all models under various patch sizes. The performance of \project is even close to the white-box attack reference on some models (e.g., Monodepth2, DepthHints and SQLDepth). Patch-RS has the second best attack performance while GenAttack performs the worse and has nearly no effect. Figure~\ref{fig:vis_main_quali} presents qualitative examples of the attack performance of different methods on PlaneDepth and FlowNetC. As shown, the first column denotes the benign scene and the MDE and OFE output. The following columns show the $2\%$ adversarial patches generated by \project and baseline methods, as well as the model output. In the first row, when the patch generated by \project is applied, the depth estimation of the patched area is significantly further than the actual distance (darker color denotes further distance estimation). In comparison, the patches generated by other methods cause less impact. In the second row, patches generated by \project and Patch-RS have degraded the OFE performance significantly, making the result unusable. More qualitative and quantitative results are in Appendix~\ref{app:more_quali}.

% \vspace{-8pt}
% \vspace{-5pt}
\subsection{Query Efficiency}
In this section, we compare the query efficiency of \project with baseline methods on the two pixel-wise regression tasks. We use DepthHints and Planedepth as the target MDE models and FlowNetC and FlowNet2 as the target OFE models. We use $2\%$ patch size and report each method's attack performance under different query times. The maximum query times are set to 300K. Figure~\ref{fig:query_effi} shows the result. As shown, on Depthhints, Planedepth and FlowNetC, \project achieves the best attack performance at various query times from the beginning. On FlowNet2, although HardBeat has a good attack performance at first, the effect is not increased with more queries. \project surpasses HardBeat at around 120K queries and causes about 19.63 EPE. Figure~\ref{fig:query_effi_001} in Appendix shows more results with 1\% patch size. In conclusion, our method is more efficient in general as it delivers superior attack performance using fewer queries to the target model. This efficiency is attributable to the more precise gradient estimation within the strategically selected square areas in \project. 

% \vspace{-5pt}
\vspace{-5pt}
\subsection{Ablation Studies}\label{sec:abl_study}

\begin{figure}[t]
     \centering
     \begin{subfigure}[b]{0.445\columnwidth}
         \centering
        \includegraphics[width=\textwidth]{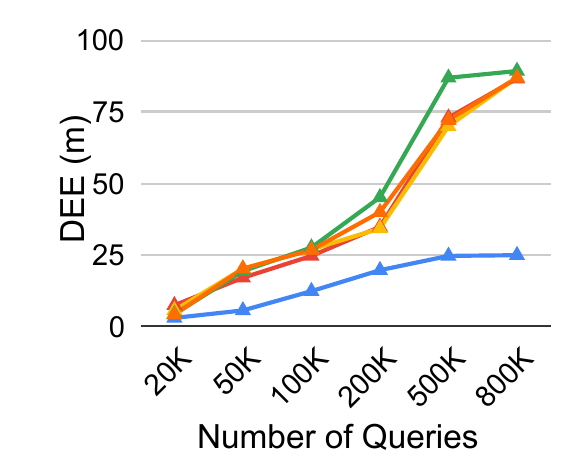}
        \caption{\small Monodepth2 }
        \label{fig:abl_trails_a}
     \end{subfigure}
     \begin{subfigure}[b]{0.535\columnwidth}
        \centering
      \includegraphics[width=\textwidth]{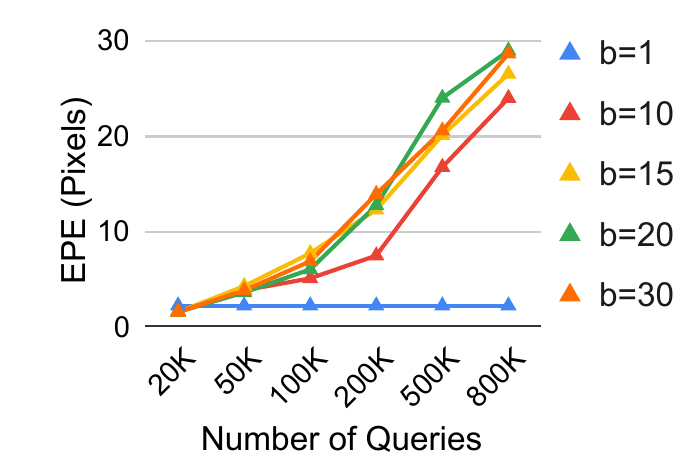}
      \caption{\small FlowNet2
      }
      \label{fig:abl_trails_b}
     \end{subfigure}
     \vspace{-10pt}
     \caption{\small Ablation study on different number of trials $b$. }
     \label{fig:abl_trails}
     \vspace{-10pt}
\end{figure}

\textbf{Number of Trials.} In Alg.~\ref{alg:GetGrad}, we leverage $b$ random noises for gradient estimation. This number of trials balances the total query times and the accuracy of the estimated gradients. We evaluate \project utilizing different numbers of trials and report the attack performance on Monodepth2 and FlowNet2 under different query times. Results are presented in Figure~\ref{fig:abl_trails}. In the two subfigures, each line denotes a choice of the number of trials $b$ used in training, and the x-axis represents the number of queries and y-axis the corresponding attack performance. As shown, less trials (e.g., $b=1$) could decrease the accuracy of gradient estimation, hence impacting the attack performance, while large trials (e.g., $b=30$) would require more query times and degrade the efficiency. $b=20$ achieves a good balance in our study and is utilized as the default settings.

\begin{figure}[t]
     \centering
     \begin{subfigure}[b]{0.43\columnwidth}
         \centering
        \includegraphics[width=\textwidth]{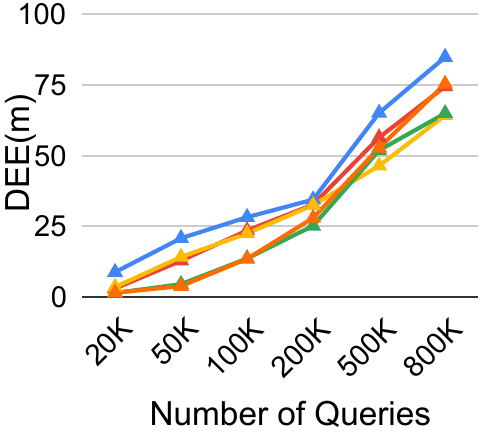}
        \caption{\small Monodepth2 }
        \label{fig:abl_intra_thres_a}
     \end{subfigure}
     \begin{subfigure}[b]{0.545\columnwidth}
        \centering
      \includegraphics[width=\textwidth]{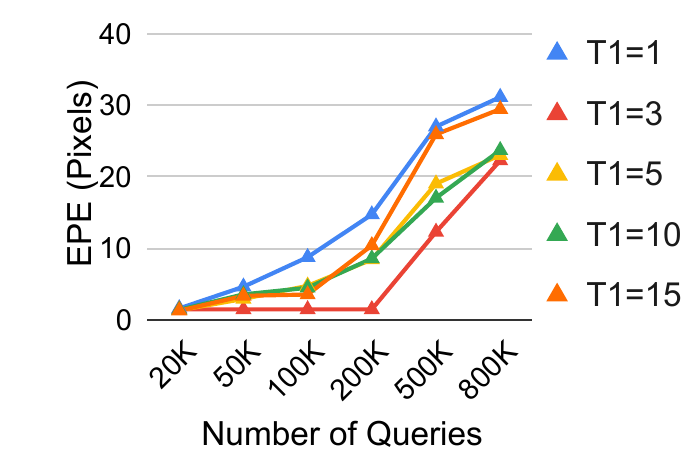}
      \caption{\small FlowNet2
      }
      \label{fig:abl_intra_thres_b}
     \end{subfigure}
     \vspace{-10pt}
     \caption{Ablation study on different intra-square threshold $T_1$.}
     \label{fig:abl_intra_thre}
     \vspace{-10pt}
\end{figure}

\textbf{Intra-square Threshold.} The intra-square threshold $T_1$ in Alg.~\ref{alg:BadPartAttack} (line 21) controls the tolerance for negative update steps within the square area. Upon reaching this threshold, a different square location will be chosen. We have adjusted the threshold, ranging from 1 to 15, to evaluate the attack performance on Monodepth2 and FlowNet2 under various query times. Figure~\ref{fig:abl_intra_thre} presents the result. As shown, \project yields optimal performance on both models when $T_1$ is set to 1, and it is adopted as our default setting. Further ablation studies concerning the hyper-parameter $T_2$ and the locations of the patch can be found in Appendix~\ref{app:ablations}.

\textbf{Design Choices.} We also conduct ablation studies to investigate the impact of our design choices. As shown in Figure~\ref{fig:overview}, our method incorporates innovative designs of probabilistic sampling (PS), score normalization (SN) and adaptive scaling (AS). We assess various combinations of these designs and report the attack performance on Monodepth2 and FlowNet2 with 300K query times. Results are shown in Table~\ref{tab:abl_design}. As shown, the integration of all three design choices yielded the best attack performance for both models. When considering each design individually, PS makes the most significant contribution and delivers the second-best performance. The other two designs can also enhance the performance to some extent. In summary, each of our unique designs plays a vital role in \project, with PS providing the most significant boost to performance.

\begin{table}[t]
    \centering
    \caption{Ablation study on different design choices.}
    \label{tab:abl_design}
    \scalebox{0.9}{
    \begin{threeparttable}
    \begin{tabular}{ccc|cc}
    \toprule
SN & AS & PS & Monodepth2 & FlowNet2 \\ \midrule\midrule
 &  &  & 38.41 & 6.90 \\
 &  & $\checkmark$ & \underline{54.88} & \underline{15.19} \\
 & $\checkmark$ &  & 43.86 & 3.39 \\
$\checkmark$ &  &  & 41.96 & 8.75 \\
$\checkmark$ & $\checkmark$ &  & 52.28 & 4.89 \\
$\checkmark$ & $\checkmark$ & $\checkmark$ & \textbf{60.46} & \textbf{17.13} \\ \bottomrule
\end{tabular}
\begin{tablenotes}
\footnotesize
    \item *~SN: Score Normalization, AS: Adaptive Scaling, PS: Probabilistic Sampling.
\end{tablenotes}
\end{threeparttable}
}
\vspace{-10pt}
\end{table}

\begin{figure}[t]
    \centering
    \includegraphics[width=\columnwidth]{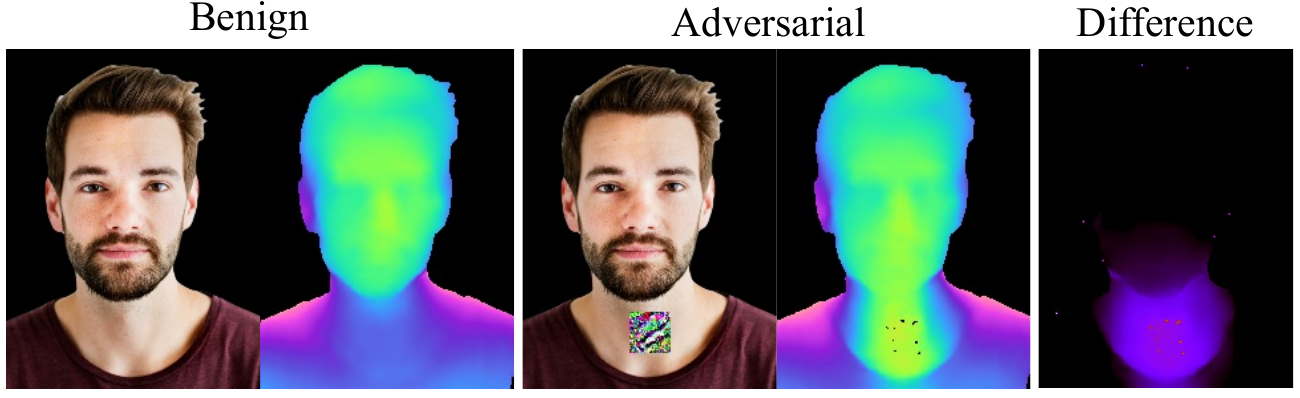}
    \vspace{-20pt}
    \caption{Attack Google MDE API~\cite{GoogleDepthApi}.}
    \label{fig:google_api}
    \vspace{-10pt}
\end{figure}

% \vspace{-8pt}
\vspace{-5pt}
\subsection{Attack Online Service}\label{sec:online}

To evaluate the feasibility of \project in real-world applications, we conduct attacks on the Google API for portrait depth estimation~\cite{GoogleDepthApi}. Note that the model is not deployed by authors and we only query it for depth estimation. We utilize ten $256\times 192$ portrait images as training set and generate a $31 \times 31$ patch using 50K queries to the API. Our attack goal is to minimize the estimated distance of the patched area on the portrait. \project has successfully reduced the mean depth estimation of the patched area by 43.5\% from 0.431 cm to 0.243 cm. Figure~\ref{fig:google_api} shows qualitative results. This adversarial example is available in the code repository, which can be uploaded to \cite{GoogleDepthApi} for efficacy validation. The adversarial patch can also be employed for beneficial purposes, such as privacy protection. By attaching the patch to personal images before publishing, individuals can prevent unauthorized use of their photos in such video composition services.

\begin{table}[t]
    \centering
    \caption{Bypassing the query-based defense.}
    \label{tab:defense}
    \scalebox{0.85}{
    \begin{tabular}{c|cc|cc}
    \toprule
    & \multicolumn{2}{c}{Monodepth2} & \multicolumn{2}{c}{FlowNet2} \\
Query & DEE & Detection Rate & EPE & Detection Rate \\ 
\midrule \midrule
50K   & 20.28 & 0\% & 1.93  & 0\% \\
200K  & 36.87 & 0\% & 8.24  & 0\% \\
400K  & 73.27 & 0\% & 19.06 & 0\% \\
800K  & 88.87 & 0\% & 25.53 & 0\% \\
\bottomrule
\end{tabular}
}
\vspace{-10pt}
\end{table}

\begin{figure*}[t!]
    \centering
    \begin{minipage}{0.45\textwidth}
    \begin{table}[H]
        \centering
        \caption{Universal attack performance with lower query budgets (2\% patch size).}\label{app:low_query_uni}
        \scalebox{0.9}{
        \begin{tabular}{ccc}
        \toprule
        {Query} & \makecell{Monodepth2\\ (DEE: Meters)} & \makecell{FlowNet2 \\(EPE: Pixels)} \\\midrule\midrule
        1K & 1.923 & 2.050 \\
        5K & 2.034 & 2.265 \\
        10K & 2.392 & 2.272 \\
        20K & 5.518 & 2.621 \\
        30K & 12.299 & 2.851\\
        \bottomrule
        \end{tabular}
        }
    \end{table}
    \end{minipage}
    \hspace{20pt}
    \begin{minipage}{0.45\textwidth}
    \begin{table}[H]
        \centering
        \caption{Single-image Attack performance with lower query budgets (2\% patch size).}\label{app:low_query_single}
        \scalebox{0.9}{
        \begin{tabular}{ccc}
        \toprule
        {Query} & \makecell{Monodepth2\\ (DEE: Meters)} & \makecell{FlowNet2\\ (EPE: Pixels)} \\\midrule\midrule
        1K & 3.200 & 2.063 \\
        5K & 6.101 & 7.548 \\
        10K & 9.060 & 13.141 \\
        20K & 13.295 & 30.102 \\
        30K & 24.235 & 41.562\\
        \bottomrule
        \end{tabular}
        }
    \end{table}
    \end{minipage}
\vspace{-10pt}
\end{figure*}

% \vspace{-8pt}
\vspace{-5pt}
\subsection{Defensive Discussion}

As pioneers in the exploration of black-box adversarial patch attacks against pixel-wise regression tasks, we find ourselves in uncharted territory with no existing defense techniques specifically tailored for this context.
Nevertheless, there are defense strategies designed against black-box attacks on classification models. For example, certifiable defenses such as PatchCleanser~\cite{xiang2022patchcleanser} employ a mask to traverse all input positions, monitoring output class mutations to identify the most suspicious position.  Furthermore, universal adversarial patch detection methods (e.g., SentiNet~\cite{chou2020sentinet}) depend on the input features responsible for the predicted class to locate the patch. However, their reliance on class output renders them unsuitable for direct application to the pixel-wise regression tasks. In contrast, query-based defense techniques, designed to detect malicious queries by black-box attacks, may be more applicable to our context.  Blacklight~\cite{li2022blacklight}, a leading defense of this type, leverages the similarity among different query inputs to detect black-box attacks. Its primary strategy involves calculating the hash representation of each incoming query and identifying an adversarial query if the hash matches any previous one. Blacklight's efficacy is contingent on the similarity between two images in consecutive queries, which is a major feature of single-image black-box attacks (e.g., SquareAttack~\cite{andriushchenko2020square}). However, \project is a universal adversarial patch attack that does not depend on sample similarity, and the randomness in different samples could potentially enhance the universal effectiveness of the generated patch. Hence we add random noise on each attack sample to by-pass the defense. Additionally, the high resolution of our input images further diminishes the efficiency and efficacy of such a defense. To assess the defensive performance of Blacklight on \project, we have incorporated it in our framework and evaluated its detection rate for varying amounts of queries. The results are presented in Table \ref{tab:defense}. As shown, for both MDE and OFE tasks, the detection rate remains zero under 800K queries, while the attack performance is not affected and continues to increase with more queries. We discuss more prior defenses in Appendix~\ref{app:atk_discuss}, and the limitations and our future work in Appendix~\ref{app:limit}.

\vspace{-5pt}
\subsection{Discussion on Lower Query Budget}

Our attack requires 50K queries to cause an average depth estimation error (DEE) of 19.43 meters on Monodepth2, which employs a U-Net architecture with two ResNet18s. This is comparable to the SOTA universal black-box attack (HardBeat~\cite{tao2023hard}),  which requires 50K queries to achieve over 80\% attack success rate on CIFAR-10 using a ResNet18 classifier. Other patch attacks requiring a smaller number of queries are for classification models, not pixel-wise regression models we are targeting, and they are not universal attacks. In particular, our scenario is characterized by the absence of class labels, precluding our ability to employ an image from the target class (e.g., an elephant) as a form of robust prior knowledge to initialize the patch on a source image (e.g., a clock), as did in DevoPatch~\cite{chen2023query}. Additionally, we argue the one-time cost of 50K queries is affordable because it could be completed within 14 hours on the platform with a rate limit of 1 query per second (e.g., ClipDrop~\cite{ClipdropApi}), and even faster on Google 3DPortrait API~\cite{GoogleDepthApi} without rate limitations. Actually, it only takes us less than 4 hours to attack this real-world API with 50K queries.

Please note that our method is an universal attack (rather than single-image attack), which is more challenging and practical. Although the patch generation process could require more queries, it is a one-time effort and the generated patch can attack arbitrary unseen images without further queries. As stated in the Attack Settings of Section~\ref{sec:eval_setup}, the attack performance of patches in our experiments is evaluated on an unseen test set, and Figure~\ref{fig:vis_appendix_scens} shows the qualitative results, demonstrating the universal effectiveness of our patches across various scenes.

It is also worth mentioning that, unlike the classification tasks in which the attacks are either successful (i.e., output a wrong label) or unsuccessful (i.e., output the correct label), for regression models, the error caused by the attack (i.e., attack performance) increases continuously with the query times. That means, lower query budgets can still cause some amount of error. In table~\ref{app:low_query_uni}, we report the attack performance of our approach under lower query budgets. It is the same universal attack setting as our main experiments. As shown, using  a lower query budget (e.g., 30K) could still cause noticeable errors on certain target models. 

If the attacker aims on a single-image attack, he can easily adapt our method to such a scenario to reduce query times (just make the training set and the validation set only contain the target image). We conduct additional experiments to evaluate the attack performance of BadPart on a single image with different query budgets. Results are shown in Table~\ref{app:low_query_single}. As shown, for both the MDE and OFE tasks, the errors caused by our single-image attack are already significant at 10K queries. 

\vspace{-5pt}
\section{Conclusion}
% \vspace{-5pt}
We propose \project, the first unified black-box adversarial patch attack against pixel-wise regression tasks, aiming at identifying vulnerabilities in visual regression models under query-based black-box attacks. \project utilizes square-based optimization, probabilistic square sampling and score-based gradient estimation, 
overcoming the scalability issues faced by previous black-box patch attacks. 
On 7 models across 2 typical pixel-wise regression tasks, our experiments compare \project with 3 baseline attack methods, validating the efficacy and efficiency of our method.

\section*{Impact Statement}

The unified adversarial patch attack that we proposed against pixel-wise regression models aims to disclose the vulnerabilities in such models under query-based black-box attacks. Our work highlights potential security risks in applications that rely on those models, such as autonomous driving, virtual reality, and video compositions. We hope to draw the attention of the related developers, and motivate the machine learning (ML) community to create more robust models or defense mechanisms against these types of attacks. This study around the robustness of models is aligned with many prior works/attacks in the ML community, and aims to advance the field of ML. Nevertheless, it is also worth mentioning that our technique can be used for benign purposes, such as protecting privacy, as we discussed in \S\ref{sec:online}.

\bibliography{example_paper}
\bibliographystyle{icml2024}

%%%%%%%%%%%%%%%%%%%%%%%%%%%%%%%%%%%%%%%%%%%%%%%%%%%%%%%%%%%%%%%%%%%%%%%%%%%%%%%
%%%%%%%%%%%%%%%%%%%%%%%%%%%%%%%%%%%%%%%%%%%%%%%%%%%%%%%%%%%%%%%%%%%%%%%%%%%%%%%
% APPENDIX
%%%%%%%%%%%%%%%%%%%%%%%%%%%%%%%%%%%%%%%%%%%%%%%%%%%%%%%%%%%%%%%%%%%%%%%%%%%%%%%
%%%%%%%%%%%%%%%%%%%%%%%%%%%%%%%%%%%%%%%%%%%%%%%%%%%%%%%%%%%%%%%%%%%%%%%%%%%%%%%
\newpage
\appendix
\onecolumn

\begin{center}
 \textbf{\LARGE Appendix}

\end{center}

\section{Experimental Details}\label{app:exp_setting}

\textbf{Attack Settings.} 
Adversarial patches are generated utilizing a single GPU (Nvidia RTX A6000) equipped with a memory capacity of 48G, in conjunction with an Intel Xeon Silver 4214R CPU. 
The resolution of input scenes from the KITTI dataset is resized to $384\times1280$ for Planedepth and $320\times1024$ for other models. We establish the initial square area as 2.5\% of the patch area, and the pre-defined square size schedule (Algorithm \ref{alg:square_sample} line 4) is set at $100, 500, 1500, 3000, 5000, 10000$ for a maximum of 10000 iterations. The square area is halved once the iteration index reaches the pre-defined steps. The initial noise bound $\alpha$ (Algorithm \ref{alg:BadPartAttack} line 7) and noise decay factor $\gamma$ (Algorithm \ref{alg:BadPartAttack} line 23) are set to $0.1$ and $0.98$ respectively. The initialization period $K$ (Algorithm \ref{alg:square_sample} line 5) is $100$ iterations. We adopt an Adam optimizer with the learning rate of $0.1$, and set $0.5$ for both $\beta_1$ and $\beta_2$. The reference white-box attack in Table~\ref{tab:atk_perf} also employs the same Adam optimizer, while the gradients for the patch region are calculated through back-propagation. Other hyper-parameters are discussed in the ablation studies, in which we use $b=20$, $T_1=1$ and $T_2=1$ as the default settings. We discuss the transferability of our approach to another dataset in Appendix~\ref{app:trans_dataset}.

\textbf{Runtime Overhead.} 
Table \ref{tab:runtime} displays the time used to generate a valid universal adversarial patch after $300K$ queries for both MDE and OFE models. The patch size is $2\%$ of the input image. The first column displays the target model name. The second column denotes the attack performance and the last column reports the runtime overhead of the patch generation.

\begin{figure}[ht]
\vspace{-25pt}
    \begin{minipage}{0.49\textwidth}
        \begin{table}[H]
        \centering
        \caption{Attack performance and runtime overhead at 300K queries.}
        \begin{tabular}{ccc}
        \toprule
             Models & DEE / EPE  & Runtime \\
             \midrule
             \midrule
             Monodepth2& 74.71  & 0.5 h\\
             Depthhints& 38.54  & 0.5 h\\
             Planedepth& 21.03  & 4 h\\
             SQLdepth  & 41.62  & 4 h\\
             \midrule
             FlowNetC  & 695.55 & 0.5 h\\
             FlowNet2  & 19.63  & 1 h\\
             PWC-Net   & 3.81   & 2 h\\
             % RAFT      &  & \\
             \bottomrule
        \end{tabular}
        \label{tab:runtime}
    \end{table}
    \end{minipage}
    \begin{minipage}{0.49\textwidth}
            \begin{table}[H]
            \vspace{-27pt}
        \centering
        \caption{Attack performance on different patch locations.}
        \label{tab:position}
        \begin{tabular}{c|cc}
        \toprule
    Query & DepthHints & FlowNetC \\ 
    \midrule \midrule
    50K   & 2.85  & 4.96   \\
    200K  & 20.99 & 128.39 \\
    400K  & 33.15 & 315.30 \\
    % 600K  & 37.91 & 416.12 \\
    800K  & 42.63 & 447.57 \\
    1000K & 46.56 & 483.54 \\
    \bottomrule
    \end{tabular}
    \end{table}
    \end{minipage}
    \vspace{-10pt}
\end{figure}

\section{Additional Ablation Studies}\label{app:ablations}

\begin{figure}[ht]
     \centering
     \begin{subfigure}[b]{0.29\columnwidth}
         \centering
        \includegraphics[width=\textwidth]{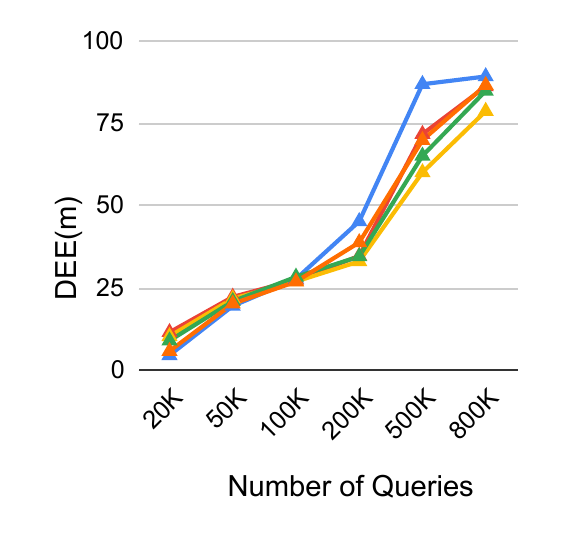}
        \caption{\small Monodepth2 }
        \label{fig:abl_inter_thres_a}
     \end{subfigure}
     \begin{subfigure}[b]{0.35\columnwidth}
        \centering
      \includegraphics[width=\textwidth]{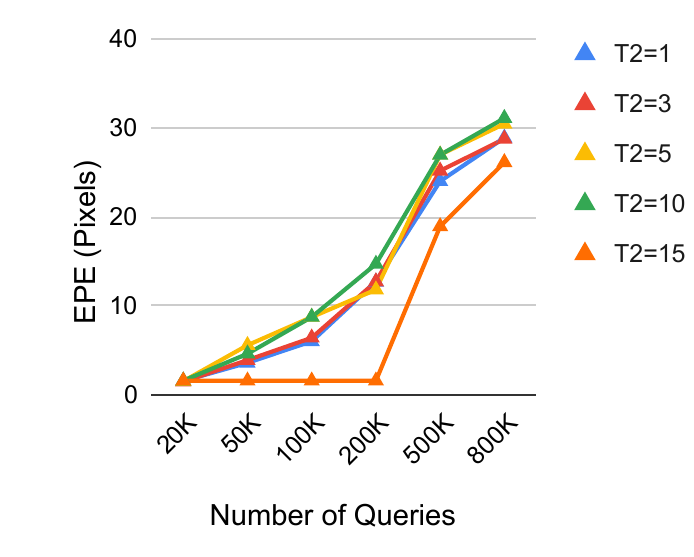}
      \caption{\small FlowNet2
      }
      \label{fig:abl_inter_thres_b}
     \end{subfigure}
     \vspace{-10pt}
     \caption{Ablation study on different inter-square threshold $T_2$.}
     \label{fig:abl_inter_thre}
     \vspace{-5pt}
\end{figure}

\textbf{Inter-square Threshold.} The inter-square threshold $T_2$ in Algorithm~\ref{alg:BadPartAttack} (line 23) controls the tolerance for negative iterations of square location selection. Upon reaching this threshold, \project reduces the noise bound $\epsilon$ for those trials in gradient estimation. Figure~\ref{fig:abl_inter_thre} presents the results of our experimental ablations on this parameter. As shown, its influence on the attack performance is not substantial, except for a large value setting on FlowNet2 (e.g., $T_2=15$). Consequently, we have set $T_2$ to 1 in our main experiments.

\begin{figure}[!t]
    \centering
    \includegraphics[width=\textwidth]{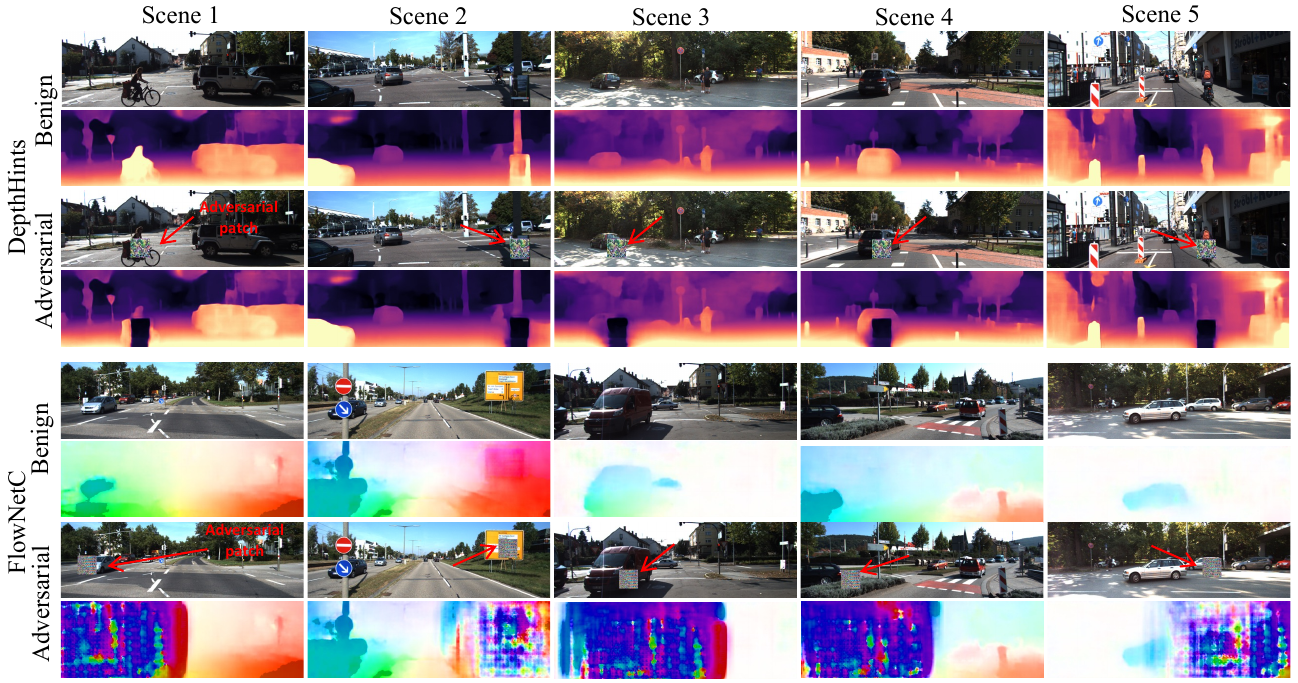}
    \vspace{-10pt}
    \caption{Qualitative examples of the attack performance for different patch locations.}
    \label{fig:vis_positions}
    \vspace{-10pt}
\end{figure}

\textbf{Patch Locations.} In consideration of different patch locations, \project can be easily extended to generate not only a scene-independent but also location-independent adversarial patch. 
For every step in optimizing the square area of adversarial patch in Algorithm~\ref{alg:BadPartAttack}(line 14-22), we randomly attach the adversarial patch on $K$ different positions ${\mathbf{q_1},...,\mathbf{q_k}}$. For each position $\mathbf{q_i}$, we get the estimated gradient $\mathbf{g_i}$ by Algorithm~\ref{alg:GetGrad}. The final gradient $\mathbf{g}$ is the average of ${\mathbf{g_1},...,\mathbf{g_k}}$. Then the current square area of adversarial patch is optimized by the estimated gradient $\mathbf{g}$. In our experiment, we utilize Depthhints and FlowNetC as our target models and the number of patch positions $K$ is set to 3. During the training stage, we randomly sample 5 patch locations on the validation images $[\mathbf{x^v}]_n$. In testing, we evaluate the attack performance on 100 random patch locations in the test set.
Other settings remain the same as the previous experiments. Table~\ref{tab:position} shows the result. We report the attack performance on two models, DepthHints and FlowNetC, under various query times. 
The average DEE/EPE caused by the adversarial patch on 100 random positions continues growing with queries rising after 1000K queries. Some qualitative examples are shown in Figure~\ref{fig:vis_positions}, using the converged patch. Columns represent various scenes. Each row in two Figures denotes a input-output pair of the target MDE/OFE model. The first two rows belong to Depthhints while the last two rows belong to FlowNetC. 
We can see that the adversarial patch generated by \project leads to significant error universally across both varying scenes and patch locations, which suggests that the patch exhibits robust characteristics of scene and location invariance.

\begin{figure}[!t]
    \centering
    \includegraphics[width=\textwidth]{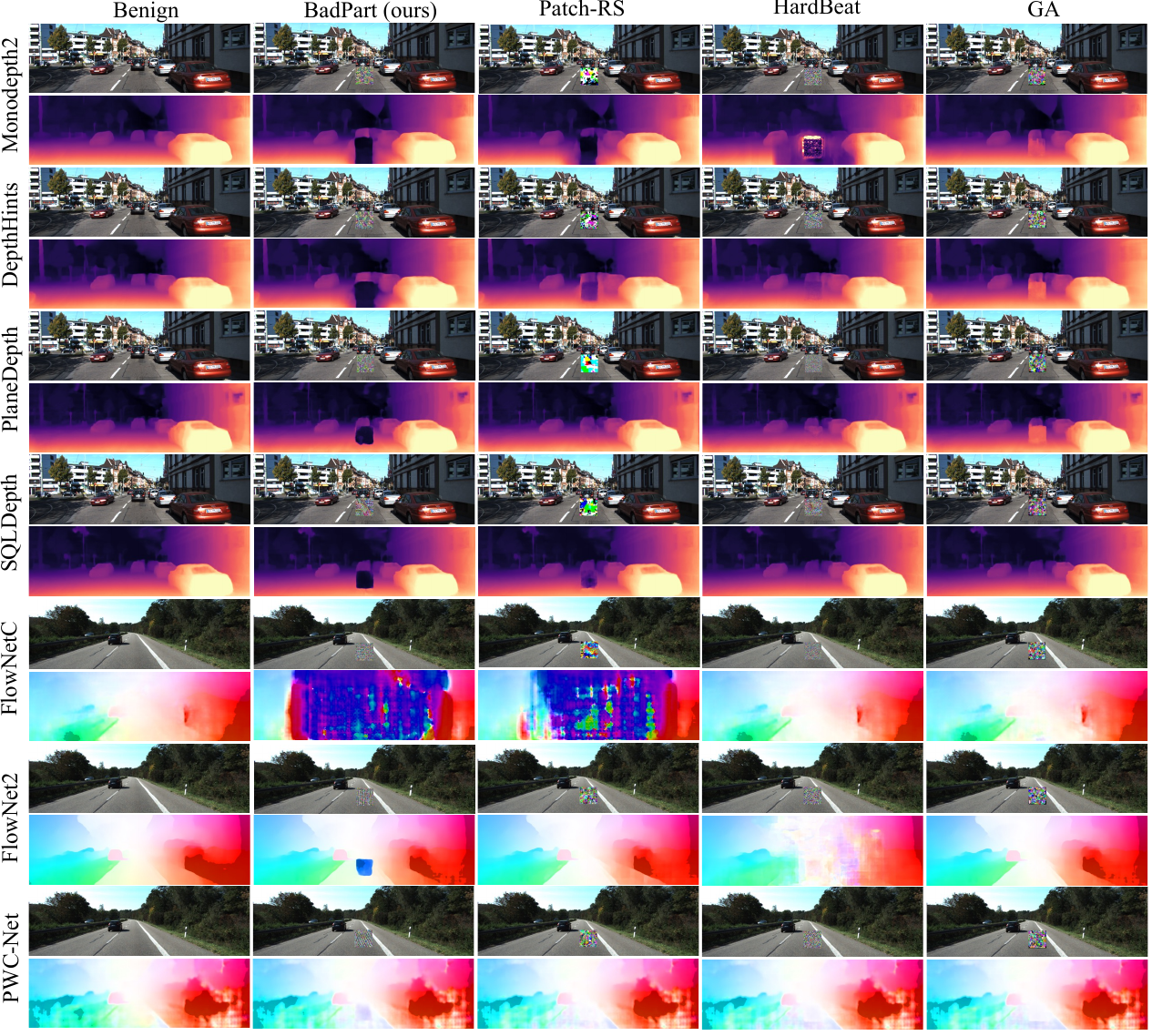}
    \caption{More qualitative examples of the attack performance of \project and the baseline methods on different models.}
    \label{fig:vis_appendix_methods}
    % \vspace{-20pt}
\end{figure}

\begin{table}[t]
\centering
\caption{More quantitative results of attack performance at 300K query times.}\label{tab:more_results}
\scalebox{0.95}{
\begin{tabular}{c|cccc|cccc|cccc}
\toprule
 & \multicolumn{4}{c}{57 * 57 Patch (1 \%)} & \multicolumn{4}{c}{80 * 80 Patch (2 \%)} & \multicolumn{4}{c}{100 * 100 Patch (3 \%)} \\
Models & GA & HB & P-RS & \textbf{Ours} & GA & HB & P-RS & \textbf{Ours} & GA & HB & P-RS & \textbf{Ours} \\ \midrule\midrule
Monodepth2 & -0.03 & 3.62 & \textbf{36.29} & 20.37 & -0.17 & 19.64 & 57.72 & \textbf{66.21} & 0.61 & 18.03 & 77.33 & \textbf{81.98} \\
DepthHints & 0.30 & 2.20 & 1.63 & \textbf{39.11} & -0.76 & 3.50 & 5.65 & \textbf{38.54} & -1.47 & 2.24 & 40.24 & \textbf{42.40} \\
SQLDepth & -0.14 & -0.02 & 3.89 & \textbf{36.28} & -0.43 & 0.12 & 15.31 & \textbf{41.39} & -0.57 & 0.45 & 8.17 & \textbf{42.31} \\
PlaneDepth & -0.43 & 0.83 & 0.83 & \textbf{17.19} & -1.75 & 1.07 & 0.99 & \textbf{21.03} & -1.86 & 1.47 & 2.47 & \textbf{19.39} \\ \midrule
FlowNetC & 5.42 & 3.70 & 5.24 & \textbf{463.30} & 3.66 & 3.62 & 347.10 & \textbf{695.50} & 4.58 & 4.43 & 304.70 & \textbf{455.60} \\
FlowNet2 & 2.28 & 12.21 & 2.64 & \textbf{18.75} & 1.67 & 7.22 & 1.77 & \textbf{19.63} & 1.27 & 10.31 & 2.96 & \textbf{11.05} \\
PWC-Net & 1.93 & 2.07 & 2.35 & \textbf{3.69} & 1.73 & 1.90 & 1.66 & \textbf{3.81} & 1.43 & 1.53 & 1.44 & \textbf{2.96} \\ \bottomrule
\end{tabular}
}
\end{table}

\vspace{-5pt}
\section{More Qualitative and Quantitative Results}\label{app:more_quali}
\vspace{-5pt}

\begin{figure}[!t]
    \centering
    \includegraphics[width=\textwidth]{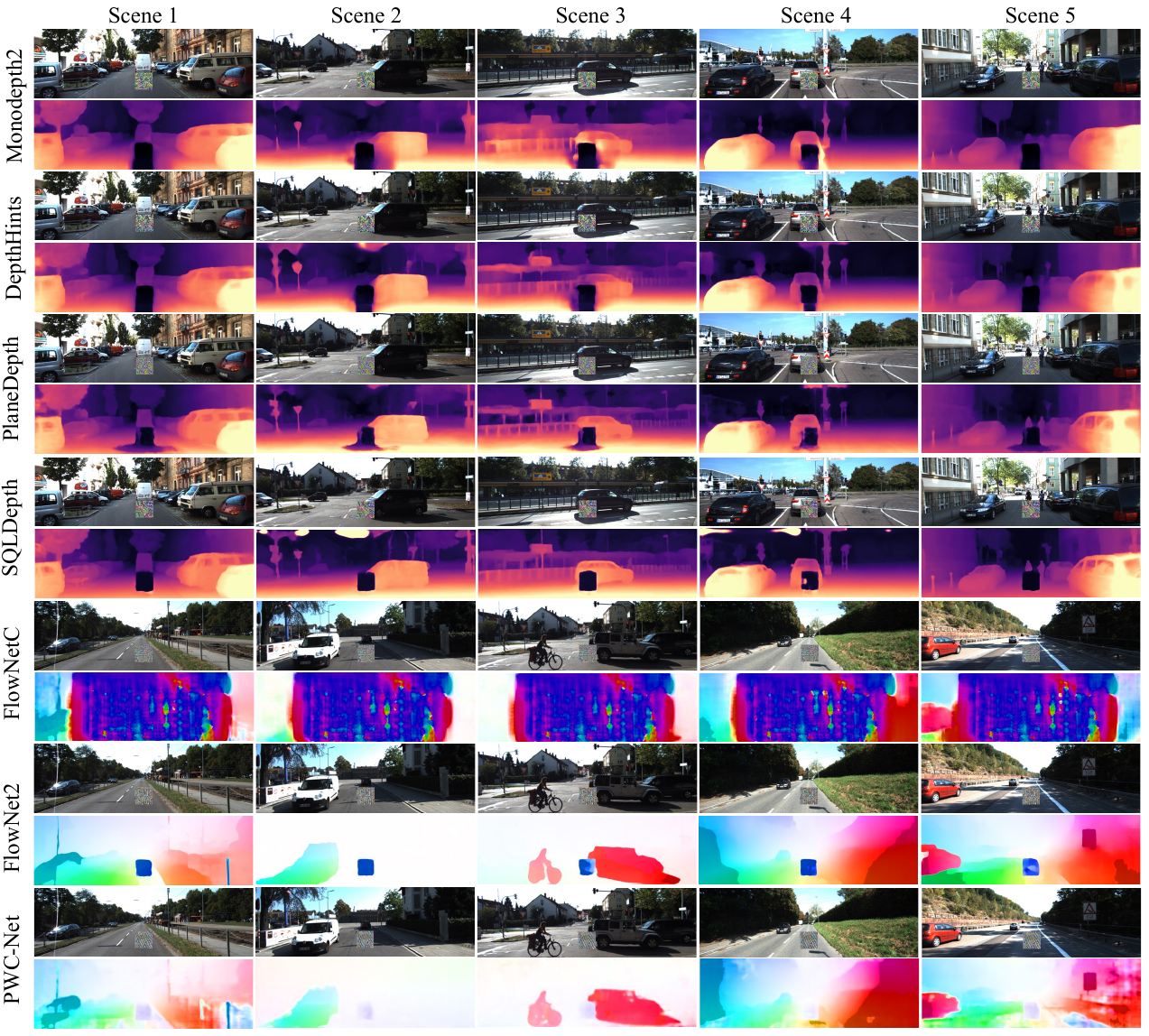}
    \vspace{-10pt}
    \caption{Qualitative examples of the attack performance of \project on different models and scene images.}
    \label{fig:vis_appendix_scens}
    \vspace{-10pt}
\end{figure}

Figure~\ref{fig:vis_appendix_methods} and Figure~\ref{fig:vis_appendix_scens} show more qualitative results of our attack. Each row in the two Figures denotes a target model. The first four rows are MDE models and the last three rows are OFE models. The columns in Figure~\ref{fig:vis_appendix_methods} denote different attack methods and the columns in Figure~\ref{fig:vis_appendix_scens} represent various scenes. For MDE models, since the attack goal is to make the distance estimation as far as possible, darker colors in the estimated depth map for the patched area refer to better attack performance. For OFE models, since the adversarial patch is attached at the same position on the two input images, the ground-truth optical flow of the patched area should be zero (i.e., white color in the flow map). Hence, in the estimated flow map, stronger colors at the patched area represent better attack performance. The patches in Figure~\ref{fig:vis_appendix_methods} are generated using 300K queries and they cover 2\% of the image size. Quantitative results are presented in Table~\ref{tab:more_results} as well as other patch sizes.  It is easy to learn from Figure~\ref{fig:vis_appendix_methods} and Table~\ref{tab:more_results} that \project has the best attack performance on various pixel-wise regression models. In addition, Figure~\ref{fig:vis_appendix_scens} shows that the generated patch works universally across varying scenes. 

\begin{figure*}[ht!]
    \centering
    \begin{subfigure}[b]{0.23\textwidth}
        \includegraphics[width=\textwidth]{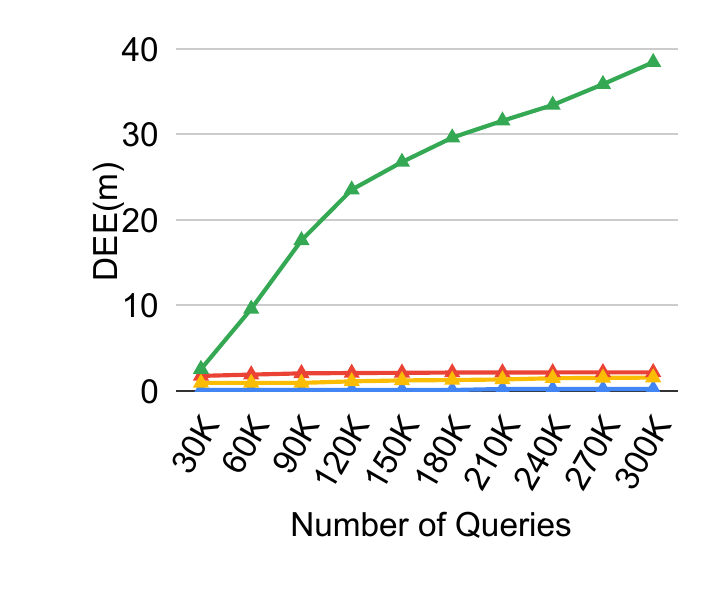}
        \caption{Depthhints}
    \end{subfigure}
    \hfill
    \begin{subfigure}[b]{0.23\textwidth}
        \includegraphics[width=\textwidth]{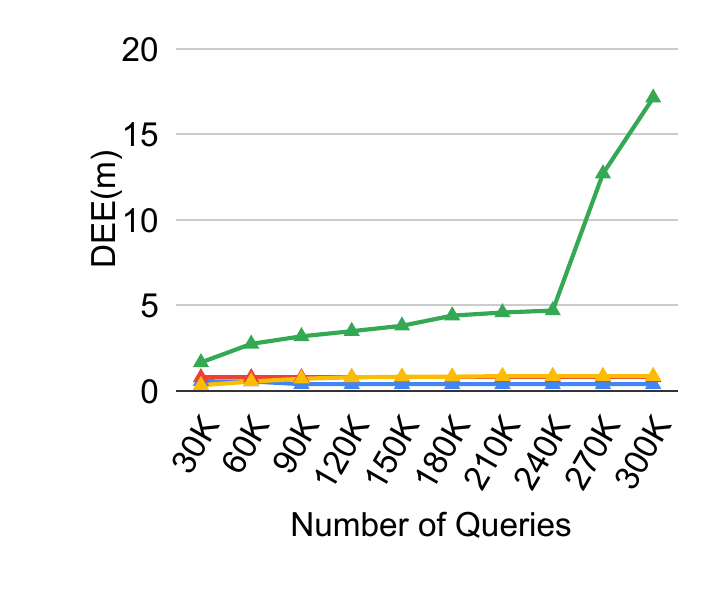}
        \caption{Planedepth}
    \end{subfigure}
    \hfill
    \begin{subfigure}[b]{0.24\textwidth}
        \includegraphics[width=\textwidth]{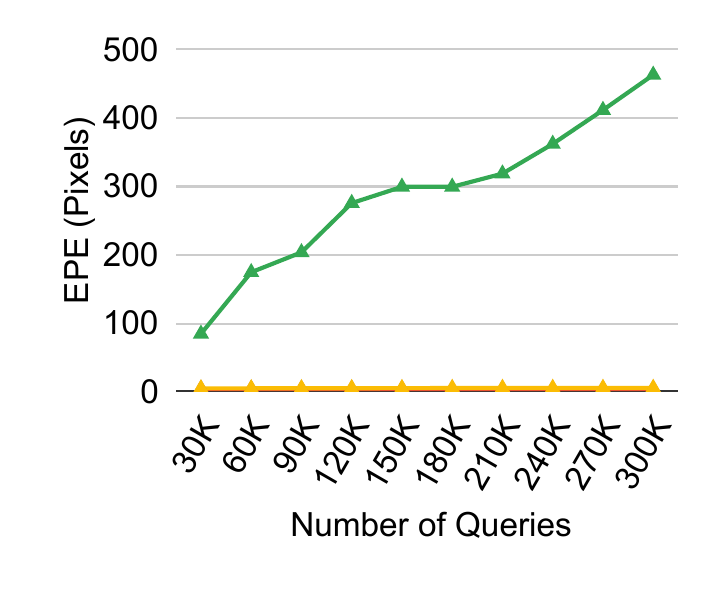}
        \caption{FlowNetC}
    \end{subfigure}
    \hfill
    \begin{subfigure}[b]{0.275\textwidth}
        \includegraphics[width=\textwidth]{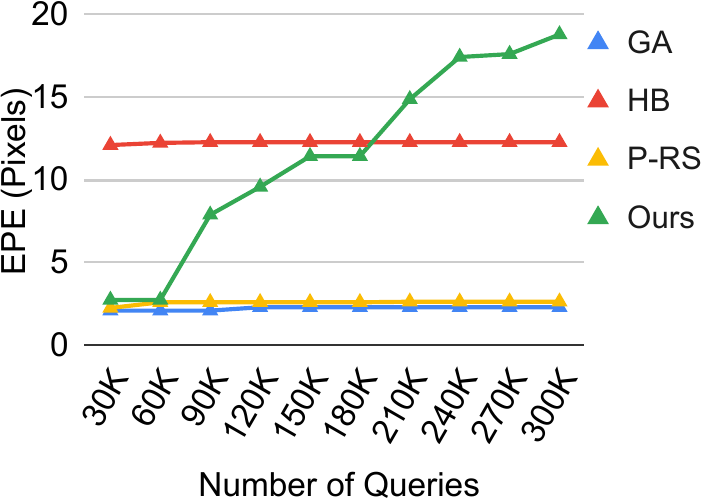}
        \caption{FlowNet2}
    \end{subfigure}
    \vspace{-10pt}
    \caption{Comparison of query efficiency between \project and the baseline methods on four models (1\% patch).}
    \label{fig:query_effi_001}
    \vspace{-10pt}
\end{figure*}

%-------------------------------------------------------------------------------------------

\section{Discussion on More Attacks and Defenses}\label{app:atk_discuss}
\textbf{Attacks.} In our main experimetns, we have endeavored to compare our method with the SOTA score-based (i.e., soft-label) attack, specifically Patch-RS~\cite{croce2022sparse}, and decision-based (i.e., hard-label) attack, namely HardBeat~\cite{tao2023hard}. Given the new challenges encountered in black-box patch attacks against pixel-wise regression models, we had to adapt these SOTA attacks which were originally designed for classification models, to ensure an equitable comparison. It is important to note, however, that not every method developed for classification models is amenable to adaptation for our  scenario. This is, in part, due to the absence of class labels in our context, which are integral to some methods.  For instance, PatchAttack~\cite{yang2020patchattack} leverages pre-generated texture images for each class in ImageNet, which are imbued with distinctive class features capable of triggering the corresponding label, generated using a white-box surrogate model. Similarly, DevoPatch~\cite{chen2023query}  utilizes an image from the target class (e.g., an elephant) as foundational prior knowledge to initiate the patch on a source image (e.g., a clock). The reliance on class labels in these methods renders them incompatible with our investigation of purely query-based black-box attacks in the realm of pixel-wise regression tasks. 

HPA~\cite{fawzi2016measuring} and Adv-watermark~\cite{jia2020adv} are also designed for classification tasks. However, HPA~\cite{fawzi2016measuring} primarily utilizes the classifier's output logits, which we can adapt to our context by using our regression outputs. Meanwhile, Adv-watermark~\cite{jia2020adv} employs an evolutionary algorithm to optimize the positioning and transparency of watermarks that serve as adversarial patches. To evaluate the attack performance of HPA and Adv-watermark on pixel-wise regression tasks, we conducted additional experiments, and results are shown in Table~\ref{tab:more_atks}. As shown, the errors caused by those attacks on both MDE and OFE tasks are limited, and additional queries do not significantly enhance their attack performance, suggesting low efficiency. The performance of Adv-watermark remains nearly static after initialization. In contrast, our approach proves to be substantially more effective and efficient.

\begin{table}[t!]
    \centering
    \caption{Attack performance comparison with additional attack baselines (single-image).}
    \label{tab:more_atks}
    \begin{tabular}{c|ccc|ccc}
    \toprule
 & \multicolumn{3}{c}{{Monodepth2}} & \multicolumn{3}{c}{{FlowNet2}} \\
{Query} & {HPA} & {Adv-watermark} & {BadPart (ours)} & {HPA} & {Adv-watermark} & {BadPart (ours)} \\
\midrule\midrule
10K & 1.376 & 2.178 & 9.060 & 3.024 & 4.551 & 13.141 \\
30K & 1.386 & 2.178 & 24.235 & 3.335 & 4.551 & 41.562 \\
50K & 1.409 & 2.178 & 50.953 & 3.542 & 4.553 & 53.663 \\
80K & 1.503 & 2.178 & 70.598 & 3.677 & 4.553 & 60.793 \\
100K & 1.598 & 2.178 & 76.890 & 3.828 & 4.554 & 67.954\\
\bottomrule
\end{tabular}
\end{table}

It is also pertinent to mention that the above attacks are all single-image attacks, in which each image requires a unique patch pattern optimized specifically. However, we focus on the universal attack, in which the patch generation is a one-time effort and the generated patch can be applied to arbitrary unseen images and attack universally. This fundamental difference in problem setting also elucidates the impracticality of applying the aforementioned methods to our scenario. 

\textbf{Defenses}. We have also further analyzed the mainstream defense methods against patch attacks and have ported the applicable defense algorithms to our scenario to test the effectiveness of our attack methods. The Local Gradient Smoothing (LGS) algorithm~\cite{naseer2019local} implements security defense through identifying the high-frequency patch areas and utilizing local gradient smoothing to degrade the patch’s effectiveness. However, in our experiments, this method results in a high rate of false positive anomaly identification and smoothing on benign areas. Although the false positive smoothing may not affect the classification tasks, it severely affected the pixel-wise regression outputs of our subject models in areas that are incorrectly smoothed. Detailed results can be found in Table~\ref{tab:more_defenses}. The algorithm's parameters are set to the optimal configuration as per the paper. As shown,  the defense affects, on average, 25.245\% of the total area in randomly selected benign images, leading to an average relative prediction error of 23.344\% across various pixel-wise regression models. This renders the defense impractical for real-world application.

\begin{table}[t!]
    \centering
    \caption{Model performance degradation on benign images caused by LGS~\cite{naseer2019local}.}
    \label{tab:more_defenses}
    \begin{tabular}{ccc}
    \toprule
{Models} & {Affected Portion (\%)} & {Relative Error (\%)} \\
\midrule\midrule
Monodepth2 & 27.335 & 16.698 \\
DepthHints & 25.944 & 17.010 \\
PlaneDepth & 31.781 & 15.738 \\
SQLDepth & 35.982 & 18.401 \\
FlowNetC & 19.037 & 29.740 \\
FlowNet2 & 19.695 & 27.507 \\
PWC-Net & 16.939 & 38.316 \\ \midrule
\textbf{Average} & \textbf{25.245} & \textbf{23.344}\\
\bottomrule
\end{tabular}
\end{table}

Moreover, many defense methods against patch attacks for classification tasks are not applicable in our scenario of pixel-wise regression tasks. DW~\cite{hayes2018visible} treats defense as a process similar to watermark removal targeted at patches. Although the guided back-propagation method mentioned in this article for constructing saliency maps of images might help in effectively identifying patch areas, the algorithm relies on the use of predicted labels, which do not exist in our scenario. RS~\cite{levine2020randomized} defends against patch attacks on classification models effectively using statistical predictions on a small number of image pixels. However, this practice is based on the fact that classification models could still predict the right label using only a fraction of the image. In contrast, pixel-wise regression models cannot make accurate predictions on partial areas; they need to process the entire scene, which makes this defense strategy not applicable. Based on our analysis and experiments on defense algorithms against patch attacks, there currently exists no defense method against our method that can balance good defense effectiveness with small impact on the performance of benign samples. The protection of pixel-wise regression models against attacks remains a significant challenge.

\section{Transferability of the Dataset}\label{app:trans_dataset}

The pre-training process in Algorithm~\ref{alg:BadPartAttack} is different from model training and we do not require access to the model’s training images. Our attack is the universal patch attack instead of single-image attack. It requires a one-time patch generation process (Algorithm~\ref{alg:BadPartAttack}), referred to as the “pre-training stage”, then the generated patch can be applied to arbitrary unseen images to attack the target model. As stated in the Attack Settings of Section~\ref{sec:eval_setup}, the attack performance in our experiments is evaluated using an unseen test set  (also customizable), and Figure~\ref{fig:vis_appendix_scens} shows the qualitative results, demonstrating the universal effectiveness of our patches across various scenes. Therefore, although we call the patch generation process in Algorithm~\ref{alg:BadPartAttack} as a “pre-training stage”, the training and validation images used in this stage are customizable. We have also conducted additional experiments in which we use images from another public dataset, named nuScenes, as our training set for patch generation, to attack models trained on the KITTI dataset. Results are shown in Table~\ref{tab:gen_dataset_trans}. As shown, patches generated using the nuScenes dataset still achieve a good attack performance, on par with  those generated using the KITTI dataset. This validates that access to the training images of the model is not a prerequisite.

\begin{table}[t!]
    \centering
    \caption{Attack performance of patches generated using different dataset.}
    \label{tab:gen_dataset_trans}
    \begin{tabular}{c|cc|cc}
    \toprule
{Model} & \multicolumn{2}{c}{{Monodepth2}} & \multicolumn{2}{c}{{FlowNetC}} \\
{Query \textbackslash Generation Dataset} & {nuScenes} & {KITTI (ref)} & {nuScenes} & {KITTI (ref)} \\\midrule\midrule
50K & 28.568 & 19.439 & 149.397 & 165.549 \\
100K & 42.541 & 28.397 & 208.646 & 270.407 \\
200K & 65.092 & 45.208 & 313.976 & 492.197 \\
300K & 78.380 & 66.210 & 503.938 & 695.503 \\
\bottomrule
\end{tabular}
\end{table}

Since our attack is dataset-independent and does not require access to the model’s training set, as long as the subject model works on the input image, the generated patch can be effective. We have conducted additional experiments to validate the transferability of the patch generated on KITTI to nuScenes. We sample 20 images randomly from the nuScenes dataset as the test set, and report the attack performance for different models using patches generated with various numbers of queries. The result can be found in Table~\ref{tab:test_dataset_trans}. As shown, the attack performance on images from the KITTI dataset is similar to the attack performance on the nuScenes dataset, which validates that our attack is dataset-agnostic.

\begin{table}[t!]
    \centering
    \caption{Attack performance of transferring the patch generated from KITTI to nuScenes dataset.}
    \label{tab:test_dataset_trans}
    \begin{tabular}{c|cc|cc|cc}
    \toprule
{Model} & \multicolumn{2}{c}{{Monodepth2}} & \multicolumn{2}{c}{{DepthHints}} & \multicolumn{2}{c}{{FlowNetC}} \\
{Query \textbackslash Test Dataset} & {nuScenes} & {KITTI (ref)} & {nuScenes} & {KITTI (ref)} & {nuScenes} & {KITTI (ref)} \\\midrule\midrule
50K & 20.481 & 19.439 & 8.285 & 6.803 & 164.573 & 165.549 \\
200K & 48.919 & 45.208 & 36.136 & 30.86 & 495.177 & 492.197 \\
400K & 80.714 & 81.097 & 44.276 & 42.832 & 753.558 & 742.817 \\
800K & 85.945 & 89.321 & 66.031 & 60.678 & 1149.475 & 1132.284\\
\bottomrule
\end{tabular}
\end{table}

\section{Limitations and Future Work}\label{app:limit}

In this work, we have explored the black-box adversarial patch attack against pixel-wise regression models, which reveals the potential vulnerabilities in such models and their expanded applications. We have addressed the domain-specific challenge of high-resolution patch optimization, and our proposed method has shown an attack efficacy that surpasses that of various established benchmarks. It also appears to be robust against state-of-the-art black-box defense mechanisms. However, it is important to acknowledge potential limitations of our study. Our focus was predominantly on digital-space attacks, wherein the perturbed pixels are utilized directly as input for the model. This approach aligns with the conventional methods employed in prior black-box attacks, as referenced in works such as \cite{andriushchenko2020square, croce2022sparse,tao2023hard}, and is deemed practical within the threat model that encompasses attacks on online services.  Nevertheless, the implications of physical-world attacks are arguably more profound, particularly in the context of autonomous driving. Physical-world attacks, such as those described in \cite{cheng2022physical,choi2022rvplayer}, necessitate a more nuanced consideration of environmental factors, including but not limited to viewing angles, distances, lighting conditions, and printing qualities. The question of how to effectively execute query-based black-box patch attacks in a physical setting remains unresolved. It is this intriguing challenge that we aim to address in our future research endeavors. An additional constraint pertains to the dimensions of the adversarial patch. Should individuals seek to employ such patches for the purpose of privacy preservation, as discussed in \S\ref{sec:online}, the current manifestation of the patch on portrait images may be conspicuously apparent. The generation of stealthy patches for pixel-wise regression tasks within a black-box context is an unresolved issue, and we earmark this as a topic for future investigation.

%%%%%%%%%%%%%%%%%%%%%%%%%%%%%%%%%%%%%%%%%%%%%%%%%%%%%%%%%%%%%%%%%%%%%%%%%%%%%%%
%%%%%%%%%%%%%%%%%%%%%%%%%%%%%%%%%%%%%%%%%%%%%%%%%%%%%%%%%%%%%%%%%%%%%%%%%%%%%%%

\end{document}